%% file: main.tex
\documentclass{article} %
\usepackage{iclr2027_conference,times}

\input{math_commands.tex}

\usepackage{hyperref}
\usepackage{url}

\usepackage[utf8]{inputenc}
\usepackage[T1]{fontenc}
\usepackage{booktabs}
\usepackage{amsmath,amssymb}
\usepackage{graphicx}
\usepackage{subcaption}
\usepackage{xcolor}
\usepackage{microtype}
\usepackage{listings}
\usepackage{xspace}
\usepackage{graphicx}
\usepackage{tikz}
\usepackage{enumitem}
\usepackage{hyperref}
\usepackage{cleveref}
\usepackage{placeins}  %
\usepackage{orcidlink}

\graphicspath{{figures/}}
\usetikzlibrary{arrows.meta,positioning,calc,shapes.geometric}

\lstdefinestyle{term}{basicstyle=\scriptsize\ttfamily,columns=fullflexible,
  keepspaces=true,breaklines=true,frame=none,xleftmargin=2pt,aboveskip=2pt,belowskip=1pt}
\newsavebox{\figpanel}

\title{Probe-and-Resume: When and How to Evaluate Agent Execution}

\newcommand{\orcid}[1]{{\hypersetup{hidelinks}\orcidlink{#1}}}
\author{%
Jiakai Xu\textsuperscript{*}\,\orcid{0009-0006-2074-9109} \quad
Tiancheng Ge\textsuperscript{*} \quad
Tujie Guan \quad
Yusheng Li \\
\textbf{Rohan Timmaraju \quad
Eugene Wu\,\orcid{0000-0003-4254-6688} \quad
Kostis Kaffes\,\orcid{0000-0002-0517-7206}} \\
Columbia University \\
{\small\texttt{\{ax2155,tg2985,tg2988,yl6009,rt2970\}@columbia.edu}} \\
{\small\texttt{\{ewu,kkaffes\}@cs.columbia.edu}}}

\newcommand{\coderepo}{\url{https://github.com/Alex-XJK/harbor-StateFork/tree/arxiv-v1}}

\newcommand{\sys}{Probe-and-Resume\xspace}

\newcommand{\Sinit}{S_0}

\newcommand{\kk}[1]{}
\newcommand{\alex}[1]{}
\newcommand{\ewu}[1]{}
\newcommand{\andy}[1]{}

\iclrfinalcopy %
\begin{document}

\maketitle
\lhead{Preprint.}
{\renewcommand{\thefootnote}{*}\footnotetext{Equal contribution.}}

\input{figures/fig_acting_loop_numbers}  %
\input{figures/tab_swepro_numbers}  %
\input{figures/tab_stronger_models_numbers}  %
\input{figures/tab_restart_numbers}  %
\begin{abstract}

Test-time search can improve agent performance by exploring alternative actions and using a critic to select promising branches. We study this approach in terminal-based computer-use environments, which expose only a partial view of a large underlying state, and find that the Monte Carlo tree search (MCTS) configuration we evaluate does not significantly improve task success. We identify two sources of poor critic performance: the agent's observations omit environment state needed for evaluation, and intermediate states offer limited evidence of whether a trajectory will succeed. At intermediate states, critics rank continuations only modestly better than chance, and environment access improves this little. We study when to invoke a critic and what access to give it to evaluate a node. Our findings motivate a policy that invokes the critic only at nodes the agent believes are final, rather than during partial progress. Further, the policy gives the critic read, write, and execute access to a fork of the computer's current and initial state, allowing it to probe and perform potentially destructive actions to surface any hidden state the trajectory omits. We call the resulting design \sys: when the agent claims to be done, the critic probes a fork of its state, and a rejected agent resumes where it stopped. On Terminal-Bench 2.1, this raises success by \LoopOverMCTSPts{} percentage points over MCTS with conventional critics, and approaches the success rate of the Codex CLI agent with the same underlying model at less than half its cost.

\end{abstract}

\suppressfloats[t]  %
\input{figures/fig_intro}
\input{sections/01_introduction}

\input{sections/02_background}

\input{sections/02_critic_analysis}
\input{sections/04_acting_critic}

\input{sections/05_conclusion}

\subsection*{AI use statement}

In this work, we used generative AI tools to provide feedback on experiment results, and implement methods.
We have not used generative AI tools to help develop theoretical models or conceptual frameworks, assist in the writing of proofs, propose or refine hypotheses,  clean and reformat dataset,  support qualitative and thematic data analysis, interpret results, 
and generate synthetic data sets, formulate mathematical claims, provide critical ingredients for proving mathematical claims, assist with translation are not applicable to this work.
Additionally, we used generative AI tools to create or modify scientific figures or images,  sourcing/searching for information,  identify relevant literature. We have reviewed all AI-assisted work. LLM-generated code was verified and tested for correctness by authors. We take responsibility for the final content of this work,
including text, claims or artifacts produced with the aid of generative AI.

\subsection*{Ethics statement}

We take research ethics seriously and have reviewed this work against the ICLR Code of Ethics. This work does not involve human subjects, personal data, or the release of a new dataset. All experiments run on the public verified benchmarks and the models we evaluate are used through their providers' APIs under their terms of service. We are not aware of ethical concerns specific to this work. We remain committed to conducting and reporting our research responsibly, and to releasing our code and logs so that the community can examine and build on it.

\subsection*{Reproducibility statement}

The framework and the semantics of its directives are given in \S\ref{sec:critic-design}, and the search policy of \sys in \S\ref{sec:acting-design}. The full text of the actor's and every critic's prompts is in our repository (\Cref{app:prompts}). The task corpus, the per-task timeouts, and the verifier annotation are listed in \Cref{app:tasks}. The statistical procedure behind every interval and paired comparison is described in \Cref{app:stats}. Configurations and the search baselines are in \Cref{app:search-baselines}, and the protocol of the reference-state critic is in \Cref{app:s0-reference}. We have made every effort to make the results reproducible. Every trial was run on a fresh AWS EC2 instance, and every number in the paper is generated by scripts from the raw trial logs. Our repository (\coderepo) provides the code, the prompts, the full configuration of every arm, and a CLI that launches each experiment and regenerates every table and figure from the logs.

\subsection*{Acknowledgments}

This research received DAPLab corporate support in the form of funding and/or compute from Amazon, IntellectAI, Infosys, Tidalwave, Veris, Shopify, Microsoft, Thinking Machines, Dandy, Perplexity, and Daytona. The views and conclusions presented here are those of the authors and should not be interpreted as representing the official positions of the funding organizations.

\bibliography{references}
\bibliographystyle{iclr2027_conference}

\clearpage
\appendix
\raggedbottom  %
\crefalias{section}{appendix}
\input{appendix/A_prompts}

\FloatBarrier
\input{appendix/B_tasks}
\FloatBarrier
\input{appendix/C_prices}
\FloatBarrier
\input{appendix/D_statistical_methodology}
\FloatBarrier
\input{appendix/E_search_baselines}
\FloatBarrier
\input{appendix/F_critic_by_position}

\FloatBarrier
\input{appendix/G_critic_cost}

\FloatBarrier
\input{appendix/H_traj_prompts}

\FloatBarrier
\input{appendix/I_s0_reference}
\FloatBarrier
\input{appendix/J_other_models}
\FloatBarrier
\input{appendix/K_restart}

\input{appendix/L_stronger_models}
\FloatBarrier
\input{appendix/M_swebench_pro}

\end{document}

%% file: math_commands.tex
\usepackage{amsmath,amsfonts,bm}

\def\eqref#1{equation~\ref{#1}}

\def\1{\bm{1}}

\DeclareMathAlphabet{\mathsfit}{\encodingdefault}{\sfdefault}{m}{sl}
\SetMathAlphabet{\mathsfit}{bold}{\encodingdefault}{\sfdefault}{bx}{n}

%% file: figures/fig_acting_loop_numbers.tex
\def\LoopNTrials{445}
\def\LoopNTasks{89}
\def\LoopRoundZero{63\%}
\def\LoopHandIn{72\%}
\def\LoopHandInPct{72\%}
\def\LoopRoundZeroPct{63\%}
\def\LoopCodexPassPct{77\%}
\def\LoopRevCap{50}
\def\LoopGainHandIn{$+$9.0 points\,[$+$5.6,\,$+$12.6]}
\def\LoopGainHandInPts{9.0}
\def\LoopOverMCTSPts{8.4}
\def\LoopMctsPassPct{64\%}
\def\LoopCostRatioMCTS{0.57}
\def\LoopCostRatioBase{2.2}
\def\LoopGapCodexClosedPct{68\%}
\def\LoopBelowCodexPts{4.2}
\def\LoopVerdictsPerTrial{1.5}
\def\LoopRejectedNotRescued{60}
\def\LoopFailSubmitted{123}
\def\LoopOfflineActFA{22\%}
\def\LoopOfflineActFR{22\%}
\def\LoopSzeroTrialsOnReq{90}
\def\LoopTasksUp{30}
\def\LoopTasksDown{1}
\def\LoopTasksSame{58}
\def\LoopNeverClaimed{27}
\def\LoopFirstPass{282}
\def\LoopFirstFail{136}
\def\LoopFailRejected{103}
\def\LoopFailAccepted{29}
\def\LoopCatchRate{76\%}
\def\LoopRescued{43}
\def\LoopRescueOfCaught{42\%}
\def\LoopRescueOfFailed{32\%}
\def\LoopPassRejected{60}
\def\LoopPassAccepted{220}
\def\LoopFailNoVerdict{4}
\def\LoopPassNoVerdict{2}
\def\LoopNoVerdict{6}
\def\LoopStoppedEarly{61}
\def\LoopFirstFR{21\%}
\def\LoopFRSurvived{57}
\def\LoopFRLost{3}
\def\LoopFRLossRate{5\%}
\def\LoopNet{40}
\def\LoopJudged{681}
\def\LoopFA{21\%}
\def\LoopFR{24\%}
\def\LoopFRLater{31\%}
\def\LoopFALater{21\%}
\def\LoopUnmetPerReject{2.4}
\def\LoopReqPerVerdict{7}
\def\LoopOneRoundShare{64\%}
\def\LoopMultiRound{149}
\def\LoopMaxRoundsHit{0}
\def\LoopCensored{0}
\def\LoopExecOverT{0.06}
\def\LoopProbeMin{1}
\def\LoopProbeCmds{4}
\def\LoopProbeSec{33}
\def\LoopClockCritic{13}
\def\LoopSzeroTrialsUsed{100}
\def\LoopSzeroTrialsProbed{412}
\def\LoopSzeroShare{24\%}
\def\LoopSzeroCopies{66}
\def\LoopCostRoundZero{\$10.70}
\def\LoopCostLoop{\$4.96}
\def\LoopCostCritic{\$7.48}
\def\LoopCostTotal{\$23.14}
\def\LoopCostCodex{\$50.29}
\def\LoopCostRatioCodex{0.46}
\def\LoopCostRatioCodexRoundZero{4.7}
\def\LoopCostCriticShare{32\%}
\def\LoopCodexPass{77\%}
\def\LoopByRoundOne{63\%}
\def\LoopByRoundTwo{70\%}
\def\LoopByRoundOnePct{63\%}
\def\LoopByRoundTwoPct{70\%}
\def\LoopByRoundThreePct{73\%}
\def\LoopByRoundThree{73\%}
\def\LoopGainByRoundTwoShare{78\%}
\def\LoopCodexPassCI{77\%\,[69\%,\,84\%]}
\def\LoopSonnetAlonePass{75\%}
\def\LoopSonnetAlonePassCI{75\%\,[69\%,\,82\%]}
\def\LoopXSonnetNTrials{444}
\def\LoopXSonnetNTasks{89}
\def\LoopXSonnetRoundZero{61\%}
\def\LoopXSonnetHandIn{70\%}
\def\LoopXSonnetHandInPct{70\%}
\def\LoopXSonnetRoundZeroPct{61\%}
\def\LoopXSonnetCodexPassPct{77\%}
\def\LoopXSonnetRevCap{50}
\def\LoopXSonnetGainHandIn{$+$9.9 points\,[$+$5.4,\,$+$14.6]}
\def\LoopXSonnetGainHandInPts{9.9}
\def\LoopXSonnetOverMCTSPts{6.5}
\def\LoopXSonnetMctsPassPct{64\%}
\def\LoopXSonnetCostRatioMCTS{13.59}
\def\LoopXSonnetCostRatioBase{38.8}
\def\LoopXSonnetGapCodexClosedPct{62\%}
\def\LoopXSonnetBelowCodexPts{6.1}
\def\LoopXSonnetVerdictsPerTrial{1.2}
\def\LoopXSonnetRejectedNotRescued{57}
\def\LoopXSonnetFailSubmitted{131}
\def\LoopXSonnetOfflineActFA{22\%}
\def\LoopXSonnetOfflineActFR{22\%}
\def\LoopXSonnetSzeroTrialsOnReq{93}
\def\LoopXSonnetTasksUp{28}
\def\LoopXSonnetTasksDown{2}
\def\LoopXSonnetTasksSame{59}
\def\LoopXSonnetNeverClaimed{27}
\def\LoopXSonnetFirstPass{269}
\def\LoopXSonnetFirstFail{148}
\def\LoopXSonnetFailRejected{106}
\def\LoopXSonnetFailAccepted{20}
\def\LoopXSonnetCatchRate{72\%}
\def\LoopXSonnetRescued{49}
\def\LoopXSonnetRescueOfCaught{46\%}
\def\LoopXSonnetRescueOfFailed{33\%}
\def\LoopXSonnetPassRejected{37}
\def\LoopXSonnetPassAccepted{222}
\def\LoopXSonnetFailNoVerdict{22}
\def\LoopXSonnetPassNoVerdict{10}
\def\LoopXSonnetNoVerdict{32}
\def\LoopXSonnetStoppedEarly{115}
\def\LoopXSonnetFirstFR{14\%}
\def\LoopXSonnetFRSurvived{32}
\def\LoopXSonnetFRLost{5}
\def\LoopXSonnetFRLossRate{14\%}
\def\LoopXSonnetNet{44}
\def\LoopXSonnetJudged{532}
\def\LoopXSonnetFA{20\%}
\def\LoopXSonnetFR{13\%}
\def\LoopXSonnetFRLater{9\%}
\def\LoopXSonnetFALater{26\%}
\def\LoopXSonnetUnmetPerReject{1.5}
\def\LoopXSonnetReqPerVerdict{6}
\def\LoopXSonnetOneRoundShare{71\%}
\def\LoopXSonnetMultiRound{122}
\def\LoopXSonnetMaxRoundsHit{0}
\def\LoopXSonnetCensored{0}
\def\LoopXSonnetExecOverT{0.06}
\def\LoopXSonnetProbeMin{2}
\def\LoopXSonnetProbeCmds{12}
\def\LoopXSonnetProbeSec{138}
\def\LoopXSonnetClockCritic{59}
\def\LoopXSonnetSzeroTrialsUsed{109}
\def\LoopXSonnetSzeroTrialsProbed{385}
\def\LoopXSonnetSzeroShare{28\%}
\def\LoopXSonnetSzeroCopies{126}
\def\LoopXSonnetCostRoundZero{\$14.31}
\def\LoopXSonnetCostLoop{\$5.41}
\def\LoopXSonnetCostCritic{\$534.67}
\def\LoopXSonnetCostTotal{\$554.38}
\def\LoopXSonnetCostCodex{\$50.29}
\def\LoopXSonnetCostRatioCodex{11.02}
\def\LoopXSonnetCostRatioCodexRoundZero{3.5}
\def\LoopXSonnetCostCriticShare{96\%}
\def\LoopXSonnetCodexPass{77\%}
\def\LoopXSonnetByRoundOne{61\%}
\def\LoopXSonnetByRoundTwo{69\%}
\def\LoopXSonnetByRoundOnePct{61\%}
\def\LoopXSonnetByRoundTwoPct{69\%}
\def\LoopXSonnetByRoundThreePct{70\%}
\def\LoopXSonnetByRoundThree{70\%}
\def\LoopXSonnetGainByRoundTwoShare{84\%}
\def\LoopXSonnetGainVsMain{$+$0.9 points\,[$-$3.4,\,$+$4.8]}
\def\LoopXSonnetHandInVsMain{$-$1.9 points\,[$-$6.7,\,$+$2.6]}
\def\LoopXGlmNTrials{444}
\def\LoopXGlmNTasks{89}
\def\LoopXGlmRoundZero{63\%}
\def\LoopXGlmHandIn{70\%}
\def\LoopXGlmHandInPct{70\%}
\def\LoopXGlmRoundZeroPct{63\%}
\def\LoopXGlmCodexPassPct{77\%}
\def\LoopXGlmRevCap{50}
\def\LoopXGlmGainHandIn{$+$7.2 points\,[$+$4.3,\,$+$10.6]}
\def\LoopXGlmGainHandInPts{7.2}
\def\LoopXGlmOverMCTSPts{6.1}
\def\LoopXGlmMctsPassPct{64\%}
\def\LoopXGlmCostRatioMCTS{2.95}
\def\LoopXGlmCostRatioBase{6.4}
\def\LoopXGlmGapCodexClosedPct{52\%}
\def\LoopXGlmBelowCodexPts{6.6}
\def\LoopXGlmVerdictsPerTrial{1.1}
\def\LoopXGlmRejectedNotRescued{43}
\def\LoopXGlmFailSubmitted{133}
\def\LoopXGlmOfflineActFA{22\%}
\def\LoopXGlmOfflineActFR{22\%}
\def\LoopXGlmSzeroTrialsOnReq{104}
\def\LoopXGlmTasksUp{22}
\def\LoopXGlmTasksDown{0}
\def\LoopXGlmTasksSame{67}
\def\LoopXGlmNeverClaimed{18}
\def\LoopXGlmFirstPass{279}
\def\LoopXGlmFirstFail{147}
\def\LoopXGlmFailRejected{76}
\def\LoopXGlmFailAccepted{58}
\def\LoopXGlmCatchRate{52\%}
\def\LoopXGlmRescued{33}
\def\LoopXGlmRescueOfCaught{43\%}
\def\LoopXGlmRescueOfFailed{22\%}
\def\LoopXGlmPassRejected{17}
\def\LoopXGlmPassAccepted{261}
\def\LoopXGlmFailNoVerdict{13}
\def\LoopXGlmPassNoVerdict{1}
\def\LoopXGlmNoVerdict{14}
\def\LoopXGlmStoppedEarly{0}
\def\LoopXGlmFirstFR{6\%}
\def\LoopXGlmFRSurvived{16}
\def\LoopXGlmFRLost{1}
\def\LoopXGlmFRLossRate{6\%}
\def\LoopXGlmNet{32}
\def\LoopXGlmJudged{507}
\def\LoopXGlmFA{44\%}
\def\LoopXGlmFR{6\%}
\def\LoopXGlmFRLater{6\%}
\def\LoopXGlmFALater{46\%}
\def\LoopXGlmUnmetPerReject{1.8}
\def\LoopXGlmReqPerVerdict{6}
\def\LoopXGlmOneRoundShare{82\%}
\def\LoopXGlmMultiRound{76}
\def\LoopXGlmMaxRoundsHit{1}
\def\LoopXGlmCensored{0}
\def\LoopXGlmExecOverT{0.05}
\def\LoopXGlmProbeMin{2}
\def\LoopXGlmProbeCmds{13}
\def\LoopXGlmProbeSec{102}
\def\LoopXGlmClockCritic{19}
\def\LoopXGlmSzeroTrialsUsed{141}
\def\LoopXGlmSzeroTrialsProbed{412}
\def\LoopXGlmSzeroShare{34\%}
\def\LoopXGlmSzeroCopies{67}
\def\LoopXGlmCostRoundZero{\$18.98}
\def\LoopXGlmCostLoop{\$3.25}
\def\LoopXGlmCostCritic{\$98.29}
\def\LoopXGlmCostTotal{\$120.52}
\def\LoopXGlmCostCodex{\$50.29}
\def\LoopXGlmCostRatioCodex{2.40}
\def\LoopXGlmCostRatioCodexRoundZero{2.7}
\def\LoopXGlmCostCriticShare{82\%}
\def\LoopXGlmCodexPass{77\%}
\def\LoopXGlmByRoundOne{63\%}
\def\LoopXGlmByRoundTwo{69\%}
\def\LoopXGlmByRoundOnePct{63\%}
\def\LoopXGlmByRoundTwoPct{69\%}
\def\LoopXGlmByRoundThreePct{70\%}
\def\LoopXGlmByRoundThree{70\%}
\def\LoopXGlmGainByRoundTwoShare{88\%}
\def\LoopXGlmGainVsMain{$-$1.8 points\,[$-$4.9,\,$+$1.6]}
\def\LoopXGlmHandInVsMain{$-$2.3 points\,[$-$6.2,\,$+$1.6]}

%% file: figures/tab_swepro_numbers.tex
\def\SweProNListed{247}
\def\SweProNHard{50}
\def\SweProNMedium{197}
\def\SweProNExcluded{20}
\def\SweProNTasks{227}
\def\SweProFirstPct{80\%}
\def\SweProSubmittedPct{88\%}
\def\SweProGain{$+$7.5 points\,[$+$3.1,\,$+$11.9]}
\def\SweProGainPts{7.5}
\def\SweProFailFirst{45}
\def\SweProCaught{36}
\def\SweProCatchRate{80\%}
\def\SweProRescued{22}
\def\SweProMissed{9}
\def\SweProPassFirst{182}
\def\SweProFalseReject{46}
\def\SweProFalseRejectRate{25\%}
\def\SweProBroke{5}
\def\SweProFalseRejectKept{41}
\def\SweProCostBase{\$6.05}
\def\SweProCostTotal{\$13.03}
\def\SweProCostRatio{2.2}
\def\SweProNVTwo{642}
\def\SweProNReleasedRuns{9}
\def\SweProNSolvedByReleased{397}
\def\SweProNNodeBB{38}
\def\SweProMediumSeed{20260924}
\def\SweProNMediumFirstDraw{60}
\def\SweProNMediumSecondDraw{137}
\def\SweProNExcludedHard{9}
\def\SweProNExcludedMedium{11}
\def\SweProNExclGo{18}
\def\SweProNExclNode{1}
\def\SweProNExclNet{1}
\def\SweProNHardScored{41}
\def\SweProNMediumScored{186}
\def\SweProNEgressRuns{391}
\def\SweProHardFirstPct{70.7\%}
\def\SweProHardSubmittedPct{73.2\%}
\def\SweProHardGain{$+$2.4\,[$-$4.9,\,$+$9.8]}
\def\SweProHardRuns{41}
\def\SweProMediumFirstPct{82.3\%}
\def\SweProMediumSubmittedPct{90.9\%}
\def\SweProMediumGain{$+$8.6\,[$+$3.8,\,$+$14.0]}
\def\SweProMediumRuns{186}
\def\SweProHardAllFirstPct{65.4\%}
\def\SweProHardAllSubmittedPct{66.3\%}
\def\SweProHardAllGain{$+$1.0\,[$-$3.9,\,$+$5.9]}
\def\SweProHardAllRuns{205}
\def\SweProFirstPctOne{80.2\%}
\def\SweProSubmittedPctOne{87.7\%}
\def\SweProGainTable{$+$7.5\,[$+$3.1,\,$+$11.9]}

%% file: figures/tab_stronger_models_numbers.tex
\newcommand{\StrongLunaLunaNTasks}{15}
\newcommand{\StrongLunaLunaNRuns}{75}
\newcommand{\StrongLunaLunaRefused}{0}
\newcommand{\StrongLunaLunaRoundZeroPct}{33\%}
\newcommand{\StrongLunaLunaHandInPct}{40\%}
\newcommand{\StrongLunaLunaGainPts}{6.7}
\newcommand{\StrongLunaLunaGainCIPts}{[$-$2.7, 16.0]}
\newcommand{\StrongLunaLunaCostRun}{\$0.05}
\newcommand{\StrongLunaLunaTimeMin}{6.7}
\newcommand{\StrongLunaLunaCostTotal}{\$4.04}
\newcommand{\StrongLunaLunaCostRefused}{\$0.00}
\newcommand{\StrongLunaLunaRunsUp}{8}
\newcommand{\StrongLunaLunaRunsDown}{3}
\newcommand{\StrongAstraFableNTasks}{15}
\newcommand{\StrongAstraFableNRuns}{75}
\newcommand{\StrongAstraFableRefused}{0}
\newcommand{\StrongAstraFableRoundZeroPct}{61\%}
\newcommand{\StrongAstraFableHandInPct}{63\%}
\newcommand{\StrongAstraFableGainPts}{1.3}
\newcommand{\StrongAstraFableGainCIPts}{[$-$6.7, 9.3]}
\newcommand{\StrongAstraFableCostRun}{\$1.69}
\newcommand{\StrongAstraFableTimeMin}{9.0}
\newcommand{\StrongAstraFableCostTotal}{\$126.94}
\newcommand{\StrongAstraFableCostRefused}{\$0.00}
\newcommand{\StrongAstraFableRunsUp}{3}
\newcommand{\StrongAstraFableRunsDown}{2}
\newcommand{\StrongAstraOpusNTasks}{15}
\newcommand{\StrongAstraOpusNRuns}{75}
\newcommand{\StrongAstraOpusRefused}{0}
\newcommand{\StrongAstraOpusRoundZeroPct}{69\%}
\newcommand{\StrongAstraOpusHandInPct}{73\%}
\newcommand{\StrongAstraOpusGainPts}{4.0}
\newcommand{\StrongAstraOpusGainCIPts}{[$-$2.7, 10.7]}
\newcommand{\StrongAstraOpusCostRun}{\$0.88}
\newcommand{\StrongAstraOpusTimeMin}{4.2}
\newcommand{\StrongAstraOpusCostTotal}{\$65.83}
\newcommand{\StrongAstraOpusCostRefused}{\$0.00}
\newcommand{\StrongAstraOpusRunsUp}{4}
\newcommand{\StrongAstraOpusRunsDown}{1}
\newcommand{\StrongOpusOpusNTasks}{15}
\newcommand{\StrongOpusOpusNRuns}{75}
\newcommand{\StrongOpusOpusRefused}{8}
\newcommand{\StrongOpusOpusRoundZeroPct}{69\%}
\newcommand{\StrongOpusOpusHandInPct}{69\%}
\newcommand{\StrongOpusOpusGainPts}{0.0}
\newcommand{\StrongOpusOpusGainCIPts}{[0.0, 0.0]}
\newcommand{\StrongOpusOpusCostRun}{\$10.61}
\newcommand{\StrongOpusOpusTimeMin}{5.4}
\newcommand{\StrongOpusOpusCostTotal}{\$52.80}
\newcommand{\StrongOpusOpusCostRefused}{\$743.28}
\newcommand{\StrongOpusOpusRunsUp}{0}
\newcommand{\StrongOpusOpusRunsDown}{0}
\newcommand{\StrongNTasks}{15}
\newcommand{\StrongAstraHubPct}{87\%}

%% file: figures/tab_restart_numbers.tex
\def\RstNRuns{439}
\def\RstNTasks{89}
\def\RstNLost{6}
\def\RstHandInPct{66\%}
\def\RstDiffHandInPts{$-$6.1}
\def\RstLossHandInPts{6.1}
\def\RstDiffHandInCIPts{[$-$10.1, $-$2.2]}
\def\RstCost{\$38.12}
\def\RstCostRatio{1.6}
\def\RstTimeMin{9.3}
\def\RstResumeTimeMin{7.8}
\def\RstRunsRestarted{161}
\def\RstRunsRestartedPct{37\%}
\def\RstRestartsTotal{320}
\def\RstRestartedOutOfTime{61}

%% file: figures/fig_intro.tex
\begin{figure}[t]
\centering
\begin{minipage}[c]{0.37\linewidth}
\centering
\begin{tikzpicture}[
  x=1.3cm,y=1cm,
  ex/.style     = {line width=0.8pt, black, -{Stealth[length=3.4pt]}},
  exL/.style    = {line width=0.8pt, black},
  probe/.style  = {line width=0.8pt, black},
  stop/.style   = {line width=1.5pt, black},
  fb/.style     = {line width=0.65pt, densely dotted, black, -{Stealth[length=3pt]}},
  st/.style     = {circle, fill=black, inner sep=0pt, minimum size=2.6pt},
  ck/.style     = {circle, draw=black, fill=white, line width=0.55pt, inner sep=0pt, minimum size=4.2pt},
  dc/.style     = {diamond, draw=black, fill=white, line width=0.7pt, inner sep=0pt, minimum size=6pt},
  dcg/.style    = {diamond, draw=black, fill=black, line width=0.7pt, inner sep=0pt, minimum size=6pt},
  vx/.style     = {black!55, font=\tiny},
  mk/.style     = {black, font=\scriptsize},
  lb/.style     = {font=\scriptsize, align=center, inner sep=1pt},
  lg/.style     = {font=\tiny, align=left, inner sep=1pt, anchor=west},
  sb/.style     = {font=\scriptsize, align=center, inner sep=1pt, black!55},
  tl/.style     = {font=\scriptsize\bfseries, align=center, inner sep=1pt},
]

\begin{scope}[shift={(0.55,2.95)}]
  \node[tl] at (0.95,1.30) {Tree search (MCTS)};
  \node[sb] at (0.95,1.05) {critic scores intermediate states};
  \node[st] at (0,0) {};
  \draw[exL] (0,0) -- (0.42,0);
  \draw[exL] (0.42,0) to[out=52,in=192] (1.00,0.34);
  \draw[exL] (0.42,0) to[out=-52,in=168] (1.00,-0.34);
  \draw[ex]  (1.00,0.34) to[out=42,in=196] (1.90,0.64);
  \draw[ex]  (1.00,0.34) to[out=-12,in=180] (1.90,0.16);
  \draw[ex]  (1.00,-0.34) to[out=12,in=180] (1.90,-0.16);
  \draw[ex]  (1.00,-0.34) to[out=-42,in=164] (1.90,-0.64);
  \node[ck] at (1.00,0.34) {}; \node[ck] at (1.00,-0.34) {};
  \node[dc] at (1.90,0.64) {}; \node[dc] at (1.90,-0.16) {}; \node[dc] at (1.90,-0.64) {};
  \node[dcg] at (1.90,0.16) {};
  \node[vx] at (0.86,0.56) {$\hat q$}; \node[vx] at (0.86,-0.58) {$\hat q$};
  \node[lb] at (0,-0.34) {$S_0$};
\end{scope}

\begin{scope}[shift={(0,0.85)}]
  \node[tl] at (1.55,1.12) {\sys};
  \node[sb] at (1.55,0.87) {critic probes a fork of candidate final states};

  \draw[exL] (0,0) -- (0.72,0);
  \foreach \x in {0,0.28} \node[st] at (\x,0) {};
  \node[dc] at (0.72,0) {};
  \node[lb] at (0.40,0.34) {$S^{(1)}$};
  \node[lb] at (0,-0.34) {$S_0$};

  \draw[probe] (0.72,0.09) to[out=58,in=180] (1.52,0.50);
  \draw[stop]  (1.52,0.38) -- (1.52,0.62);
  \node[mk] at (1.72,0.50) {$\times$};
  \draw[exL] (0.72,-0.09) to[out=-38,in=180] (1.44,-0.44);
  \draw[exL] (1.44,-0.44) -- (2.14,-0.44);
  \node[st] at (1.66,-0.44) {};
  \node[st] at (1.90,-0.44) {};
  \draw[fb] (1.50,0.30) to[out=250,in=80] (1.34,-0.36);

  \node[dcg] at (2.16,-0.44) {};
  \draw[probe] (2.16,-0.35) to[out=58,in=180] (2.90,0.02);
  \draw[stop]  (2.90,-0.10) -- (2.90,0.14);
  \node[mk] at (3.10,0.02) {\checkmark};
  \node[lb] at (2.20,-0.80) {$S^{(2)}$};
\end{scope}

\begin{scope}[shift={(0,-0.55)}]
  \node[st] at (0.05,0) {};            \node[lg] at (0.14,0) {state};
  \node[ck] at (0.05,-0.30) {};        \node[lg] at (0.14,-0.30) {search node};
  \node[dc] at (0.05,-0.60) {};        \node[lg] at (0.14,-0.60) {candidate final state};
  \node[dcg] at (1.85,0) {};           \node[lg] at (1.94,0) {submitted, graded};
  \draw[probe] (1.67,-0.30) -- (1.95,-0.30); \draw[stop] (1.95,-0.40) -- (1.95,-0.20);
                                       \node[lg] at (2.02,-0.30) {probe fork, discarded};
  \draw[fb] (1.67,-0.60) -- (1.97,-0.60); \node[lg] at (2.02,-0.60) {verdict; actor resumes};
\end{scope}
\end{tikzpicture}
\end{minipage}\hfill
\begin{minipage}[c]{0.61\linewidth}
\centering
\includegraphics[width=\linewidth]{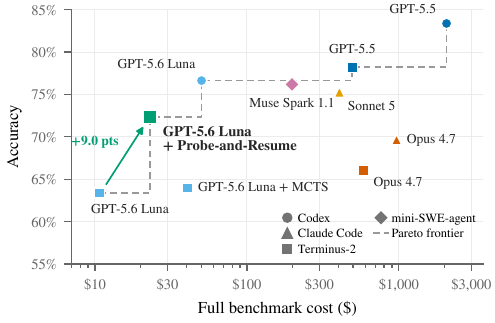}
\end{minipage}
\caption{Left: tree search scores partial states with a critic and submits a leaf; \sys calls the critic only when the actor claims to be done, lets it probe a fork of that state, and resumes the actor on a rejection. Right: accuracy against the cost of running all of Terminal-Bench~2.1 (89 tasks, five trials each); the shape is the agent harness. The GPT-5.6 Luna squares are our three configurations (no search, MCTS, \sys), priced at the AWS Bedrock rate like GPT-5.6 Luna + Codex; the other points and their costs are from Harbor Hub (\Cref{app:prices}).}
\label{fig:intro}
\end{figure}
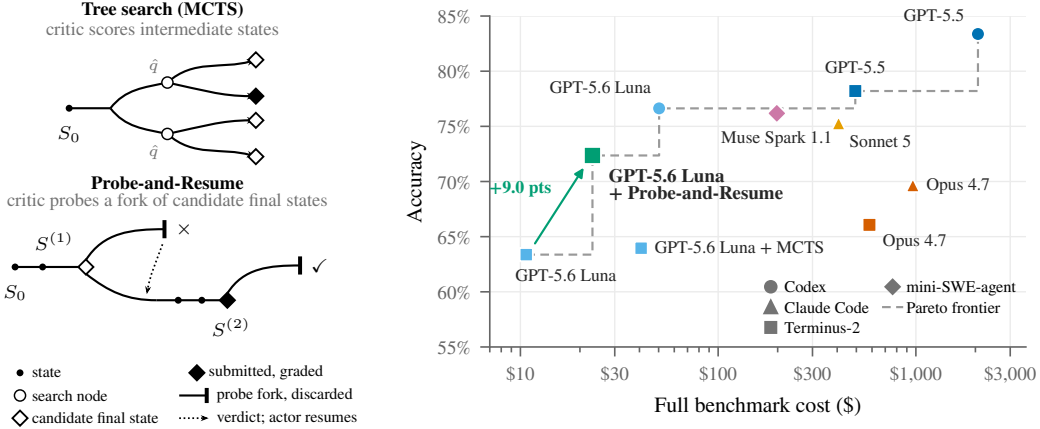

%% file: sections/01_introduction.tex
\section{Introduction}
\label{sec:intro}

\input{figures/fig_acting_loop_numbers}
Terminal agents accomplish tasks by changing a computer's state: editing files, installing packages, running programs, and configuring services. These tasks often require long sequences of actions, and a mistaken command can destroy useful state or send subsequent work down an unsuccessful path.  Recently developed branchable execution environments~\citep{xu2025foundations,dong2026deltabox,wu2026crab} allow an agent to snapshot a running environment, restore earlier states, and explore alternatives in isolated forks. This makes test-time search practical for agents.

Unfortunately, branching alone does not tell an agent which execution trajectories to pursue. Search also needs an evaluation signal: which states are promising, which require further work, and which satisfy the task?  In most automation settings, the task’s ground-truth verifier is unavailable during execution, and search must rely on a critic to evaluate candidate states to guide which executions to continue and when the task is done.   

This setting places importance on critic design, namely: when should the critic evaluate an execution, and what access does it need to produce a useful judgment?  An intermediate state asks the critic to assess its promise and whether continuing is likely to succeed, whereas a state the actor declares is done requires the critic to assess whether the task’s requirements have been met.  Both are challenging in terminal environments because the actor typically only records its reasoning, the tool calls, and the outputs it observed.  In contrast, the search state also includes that of the computer, which encapsulates files, running processes, memory, installed packages, and databases, which may be changing as side-effects of agent actions or concurrent processes, and are not visible to the actor or the critic.

We investigate these questions on Terminal-Bench~2.1~\citep{Marten_Terminal-Bench_2026}. We find that merely making branching available does not improve task success because the actor rarely uses snapshot and restore capabilities, and a Monte Carlo tree search (MCTS) search policy guided by a trajectory-only critic produces little improvement. One reason for the latter is that the critics we test are poor value estimators of intermediate states, and access to a fork of the environment helps little. However, when the actor signals completion, execute access to a fork of the environment substantially improves their ability to distinguish successful from unsuccessful outcomes. These findings motivate concentrating critic computation at candidate final states, where it can test the actor’s claim against the environment.

Our main observation is that \textbf{\it evaluating a terminal task is itself a terminal task}. Thus, neither the interaction history alone nor simply inspecting the environment is always sufficient to assess that a script, program, configuration change, or service is correct.  The critic must obtain evidence by running programs and constructing checks. Branchable environments make such evaluation, which may modify or destroy state, possible by using a disposable fork. A fork of the initial environment additionally provides a reference for requirements about what must be preserved or changed.

We therefore design \sys (\Cref{fig:intro}(left)): when the actor signals completion, the critic assesses the candidate with terminal access to forks of that state and of the initial environment. If it accepts, the candidate is submitted; otherwise the actor receives its findings and resumes from the unmodified state. Unlike MCTS, the critic judges completions, not intermediate states.

On all 89 tasks of Terminal-Bench~2.1, with five trials per task and GPT-5.6 Luna as the actor and the basic Terminus-2 harness, \sys increases success from 63.4\% at the actor’s first completion claim to 72.4\% at submission, expanding the accuracy--cost Pareto frontier without the need for new models nor new harnesses (Figure~\ref{fig:intro}(right)). The resulting accuracy is higher than Opus 4.7 using Claude Code, and over an order of magnitude cheaper. On SWE-bench Pro, the gain is similar.
Together, these results highlight the unique characteristics of critics in computer environment tasks, and how critics can turn branchable execution into useful feedback for terminal agents when ground-truth verification is unavailable.

%% file: sections/02_background.tex
\section{Background and Related Work}
\label{sec:background}

\paragraph{Test-time search and evaluation.}
Search uses additional inference compute to explore candidate solutions. Policy- and value-guided search has improved performance in games~\citep{silver2016mastering,silver2018general}, mathematical reasoning~\citep{hao2023reasoning,guan2025rstar}, and software and web tasks~\citep{zhou2024lats,koh2024treesearch,antoniades2024swesearch}. Sampling-based methods evaluate completed candidates or aggregate their answers~\citep{cobbe2021training,wang2023selfconsistency,brown2024large}; beam search, best-first search, and MCTS also use evaluations to allocate computation among partial solutions~\citep{xie2023self,yao2023tree,kocsis2006uct,coulom2006efficient,browne2012survey}. Process evaluation has proved useful in mathematical reasoning~\citep{lightman2024lets,wang2024mathshepherd}, but imperfect assessments can cause search to discard useful candidates~\citep{stroebl2026limits,yu2025scaling}. We examine intermediate predictions and completion judgments separately, because they ask different questions about an execution state.

\paragraph{Feedback and iterative refinement.}
Self-Refine uses the same model to generate an output, critique it, and revise it~\citep{madaan2023selfrefine}. Reflexion stores verbal reflections from previous trials to guide subsequent attempts~\citep{shinn2023reflexion}. Its evaluator can use environmental feedback, heuristics, model judgments, or generated tests; an evaluator and a feedback loop are therefore established components. Our method also uses feedback to revise an attempt. It preserves both the actor's environment state and trajectory when it claims the task is done, runs the critic in forks, and resumes the actor from that same candidate after rejection. This differs from resetting the environment for another trial, as in Reflexion's ALFWorld experiments~\citep{shinn2023reflexion}. %

\paragraph{External evidence for critics.}
\citet{huang2023selfcorrect} find that intrinsic self-correction, without external feedback, can fail to improve reasoning and can reduce accuracy in the settings they study. CRITIC obtains external evidence through tool interaction before revising an output~\citep{gou2023critic}, including evidence from search and code execution. Other work executes tests against candidate programs or patches~\citep{li2022competition,jain2025r2egym,ehrlich2025codemonkeys,li2025sstar}, or gives judges tools to inspect and interact with an environment~\citep{zhuge2024agentjudge,shi2026ajbench}. These methods establish the value of tool-based feedback. We study the additional choices required when the object being judged is a live environment: whether the critic should inspect intermediate or completed states, which operations it needs, and whether the current state contains the evidence required by the task. The benchmark verifier and its output are not available during search.

\paragraph{Judging agent trajectories.}
Web-agent search uses model assessments of interaction histories~\citep{koh2024treesearch}; software-agent discriminators and reward models also use trajectories and patches~\citep{antoniades2024swesearch,shum2025swerm}. On Terminal-Bench~2, pairwise judgments select among completed trials and recover part of the gap between a single attempt and oracle selection over five samples~\citep{kwok2026verifier}. Judge performance varies across web-agent benchmarks~\citep{lu2025agentrewardbench}, and execution-free patch judgments can rely on stylistic features~\citep{jain2025r2egym}. These findings motivate measuring critic errors directly, alongside task success.

\paragraph{Critics in branchable environments.}
Terminal tasks can depend on files, running processes, memory, and changes made during execution, only part of which appears in the interaction history. Revisiting a state can require replaying actions~\citep{koh2024treesearch}. Waypoint~\citep{Xu_Waypoint_2026} instead supports snapshots, restores, and isolated forks of the filesystem, processes, memory, and terminal session. We use this capability to let a critic execute checks on a disposable fork of the candidate final state, including checks that modify or destroy state. A second fork of the initial environment provides a reference for preservation and recovery requirements. Only the critic's feedback returns to the actor; its mutations are discarded, and the actor resumes from the candidate state. Our contribution combines this evaluation procedure with an empirical analysis of when critics make useful judgments and which errors arise from missing execution access or initial-state evidence (\S\ref{sec:critic-design}).

%% file: sections/02_critic_analysis.tex
\section{Critic Design for Search in Terminal Environments}
\label{sec:critic-design}

\input{figures/tab_seq_fork_numbers}
\input{figures/tab_mcts_numbers}
\input{figures/fig_critic_by_position_numbers}
\input{figures/tab_access_class_numbers}
\input{figures/fig_s0_reference_numbers}

We first compare branching tools and MCTS with the no-search baseline (\S\ref{sec:q-search}). We then examine three critic design choices: when to evaluate a state and what operations the critic may perform (\S\ref{sec:q-act}), and whether it needs the initial state as a reference (\S\ref{sec:q-info}).

\paragraph{Agent setup.}
Following policy- and value-guided search, we separate action generation, state evaluation, and search~\citep{silver2016mastering,zhou2024lats}. The actor chooses and executes actions, the critic evaluates states, and the search algorithm selects which states to continue and when to stop.
Each search node stores environment state $S_t$ and the trajectory prefix $\tau_{\le t}$ that produced it. The environment supports \textsc{snapshot}$(S)\to i$, \textsc{restore}$(i)$, and \textsc{fork}$(i)$, where $i$ identifies a saved state and the fork is an isolated live copy.

\paragraph{Experimental setup.}
We use all 89 tasks in Terminal-Bench~2.1~\citep{Marten_Terminal-Bench_2026}, run through Harbor~\citep{harbor2026} with the Terminus-2 agent and GPT-5.6 Luna. The branchable environment uses Waypoint~\citep{Xu_Waypoint_2026}, which creates, restores, and forks a Docker-like container's state.   By default, the critic uses the same model as the actor and receives the task instruction and $\tau_{\le t}$.
Once search submits a final state, we run the verifier.  Each task runs five times and we  report pass@1 with 95\% bootstrap intervals over tasks (\Cref{app:stats}). Each trial gets the task's time limit $T$ (\Cref{app:tasks}), and $T$ covers everything the trial does: the actor's turns, every critic call, and every snapshot, restore and fork. Only the verifier's grading is not charged. A trial that ends with no state to submit counts as a failure. We drop the \the\numexpr\SeqForkNLost+\MctsNLostCkpt+\MctsNNoGrade\relax{} trials that could not be graded, all from the two search baselines (\Cref{app:search-baselines}). Costs are model tokens only, priced at the AWS Bedrock rate (\Cref{app:prices}). The configuration of each search arm is in \Cref{app:search-baselines}.

\subsection{Search Baselines}
\label{sec:q-search}

We compare two search baselines. The first gives the actor \textsc{snapshot} and \textsc{restore} tools, without a separate critic. The second uses MCTS with a trajectory-only critic: it calls \textsc{snapshot} every five turns, expands at most two children, and selects nodes using UCT~\citep{kocsis2006uct}.

Table~\ref{tab:search-null} 
shows that branching tools change pass@1 from $\SeqPass\to \SeqForkPass$, while MCTS changes it $\SeqPass{} \to \MctsPass{}$ at \MctsCostRatio$\times$ the model cost.
Branching tools are rarely used, and the evaluated MCTS configuration provides little improvement:  the actor called branching tools in \SeqForkToolTrials{} of \SeqForkNTrials{} trials: it saved \SeqForkAgentSnapshots{} snapshot and restored \SeqForkRestores{} times (twice to $\Sinit$).
For instance, in the \textsc{db-wal-recovery} task, restoring $\Sinit$ recovers a write-ahead log removed after the actor opens the database. In \textsc{path-tracing}, the actor saves a state, tries a change, and restores the saved state when the change worsens the result.

These results motivate measuring critic assessments directly; they do not by themselves establish why search fails to improve performance.

\begin{table}[!ht]
\caption{Branching tools and MCTS don't improve pass@1 on Terminal-Bench 2.1  (GPT-5.6 Luna).}
\label{tab:search-null}
\vspace{4pt}
\centering
\small
\begin{tabular}{lccc}
 &\textbf{ Pass@1 }& \textbf{$\Delta$ vs.\ baseline [95\% CI] }& \textbf{Total model cost} \\
Baseline & \SeqPass & -- & \SeqCost \\
Actor + branching tools & \SeqForkPass & \SeqForkDeltaSeq\,\SeqForkDeltaSeqCI & \SeqForkCost \\
MCTS with trajectory-only critic & \MctsPass & \MctsDeltaSeq\,\MctsDeltaSeqCI & \MctsCost \\
\end{tabular}
\end{table}

\subsection{What the Critic Can Do and When to Invoke It}
\label{sec:q-act}

We now vary the critic's environment access and the intermediate state it evaluates.
We find that environment access improves the critics' predictions at intermediate states only slightly, but improves their judgments at candidate final states considerably. These results motivate concentrating execution-based assessment at candidate final states in our proposed search algorithm. These measurements do not rule out search with a stronger value estimator or another search policy, and do not show that completion is the optimal time to intervene.

\paragraph{Critic variants.}
$C^{\mathrm{traj}}$ receives no environment tools. $C^{\mathrm{read}}$ can inspect a fork of the state using read-only commands such as \texttt{cat}, \texttt{ls}, \texttt{grep}, and \texttt{diff}. $C^{\mathrm{act}}$ can execute arbitrary commands in the fork, including installing packages, writing files, running tests, and taking destructive actions (prompts in \Cref{app:prompts}). Comparing these variants distinguishes evidence in $\tau_{\le t}$ from evidence obtained by inspection or execution. The two tool-enabled critics issue commands until they return an assessment or reach their limits (\Cref{app:tasks}). %

\paragraph{Evaluation positions and metrics.}
We evaluate each critic at 25\%, 50\%, and 75\% of the actor's trajectory and at its candidate final state. At intermediate states, the critic estimates continuation success, $\hat q(S_t)$. We compare this estimate with the pass rate of \CbpSamples{} rollouts from $S_t$, using the same actor and remaining task budget. At a candidate final state, the critic returns a verdict $\hat v$, which is compared with the verifier's output. Critics do not observe the sampled continuations or verifier outcomes. We evaluate \CbpNMid{} intermediate and \CbpNFinal{} candidate final states.

Intermediate and candidate final states have different targets. We score each kind of state by the decision search makes there, and we do not compare scores between these two states (Figure~\ref{fig:critic-by-position}). At an intermediate state, search chooses which state to continue, so  the critic's ranking of states matters. We report within-task pairwise accuracy: among pairs of states from the same task and depth, the fraction that the critic orders the same way as their pass rates. A critic that guesses at random scores 50\%. Each pass rate is estimated from only \CbpSamples{} rollouts, so sampling noise also pulls the measured accuracy towards 50\% and understates the critic's ranking ability (\Cref{app:critic-by-position}). At a candidate final state, search accepts or rejects, so we report the false-reject and false-accept rates.

\input{figures/fig_critic_by_position}

\paragraph{Results by position.}
At intermediate states, the critics rank states better than chance late in the trajectory (\CbpPairLateLo--\CbpPairLateHi{} at 75\% depth), but not early (\CbpPairEarlyLo--\CbpPairEarlyHi{} at 25\%; Figure~\ref{fig:critic-by-position}a). At 25\% depth, we do not detect a statistically significant selection gain over random choice (\Cref{app:critic-by-position}). Environment access changes this little: pooled over the depths, the gain of $C^{\mathrm{act}}$ over $C^{\mathrm{traj}}$ is \CbpPairGapActTrajPooled. At candidate final states, $C^{\mathrm{act}}$ reduces the misclassification rate from \CbpReadMaeFinal--\CbpTrajMaeFinal{} to \CbpActMaeFinal. Most of the gap is in false rejects: over 50\% for $C^{\mathrm{traj}}$ and $C^{\mathrm{read}}$, around 20\% for $C^{\mathrm{act}}$ (Figure~\ref{fig:critic-by-position}b). 

\paragraph{Completion errors by verifier type.}
Table~\ref{tab:access-class} groups tasks by what their verifier uses to evaluate: reading files in the environment, executing an artifact or probing a service, or comparing against a reference value unavailable from the environment (\Cref{app:tasks}).  A false reject is a correct state incorrectly rejected, and a false accept is the opposite.

\input{figures/tab_access_class_compact}

Even when the verifier only reads files, $C^{\mathrm{read}}$ rejects \AccRecReadFR{} of passing states, compared with \AccRecActFR{} for $C^{\mathrm{act}}$. In \AccRecReadNEshare{} of these read-only false rejects, the critic reports insufficient evidence rather than a violation of a requirement. The critic may need to compute a check even when the verifier can compare the result against a known reference.

When the verifier relies on a reference value that cannot be derived from the environment,
$C^{\mathrm{act}}$ falsely accepts \AccOraActFA{} of failing states, compared to \AccExeActFA{} on tasks whose verifier executes the artifact or probes a service. The critic tends to agree with the actor's interpretation of the task completion and cannot reach the verifier's (hidden) reference.

\subsection{What Information to Give the Critic}
\label{sec:q-info}

Some requirements refer to the initial state: a file must remain unchanged, a deleted input must be recovered, a program must keep its original behavior. The candidate final state alone cannot settle them, and in \S\ref{sec:q-act} such requirements account for \AccActFRsZero{} of $C^{\mathrm{act}}$'s \AccAllActFRcount{} false rejects.
We therefore compare $C^{\mathrm{act}}$ with $C^{\mathrm{act}}{+}\Sinit$, which receives a live fork of the initial state. We rebuild the candidate final states of \SzrAllNTasks{} tasks and let each critic judge every state twice from the same snapshot (\Cref{app:s0-reference}). The two verdicts per state also measure a critic's self-consistency.

Over all \SzrAllNStates{} states (Figure~\ref{fig:s0-reference-all}), the fork raises accuracy from \SzrAllAccCtrl{} to \SzrAllAccS{} (\SzrAllAccGap). The gain comes from false rejects, which fall from \SzrAllFRCtrl{} to \SzrAllFRS{} (\SzrAllFRGap); false accepts do not move (\SzrAllFAGap), and disagreement between a critic's two verdicts on a state drops from \SzrAllFlipCtrl{} to \SzrAllFlipS{} (\SzrAllFlipGap).

\input{figures/fig_s0_reference_bars}

$C^{\mathrm{act}}{+}\Sinit$ uses the fork in \SzrAllOpenShare{} of its evaluations, on \SzrForkNStates{} states. The critic chose these states itself, so they show where the gain arises, not the effect of the fork on an average state. On them (Figure~\ref{fig:s0-reference}), accuracy rises from \SzrForkAccCtrl{} to \SzrForkAccS{} (\SzrForkAccGap) and false rejects fall from \SzrForkFRCtrl{} to \SzrForkFRS{}; on the other \SzrNoForkNStates{} states, accuracy changes by \SzrNoForkAccGap. In \textsc{modernize-scientific-stack}, for example, $C^{\mathrm{act}}$ rejects because it cannot check whether a legacy file was preserved; $C^{\mathrm{act}}{+}\Sinit$ compares the file's hash in both states and accepts. The initial state thus gives the critic more evidence for its checks: its verdicts become more accurate and consistent, not stricter.

%% file: figures/tab_seq_fork_numbers.tex
\def\SeqForkPass{61.0\%}
\def\SeqPass{63.4\%}
\def\SeqForkDeltaSeq{$-2.4$}
\def\SeqForkDeltaSeqCI{[$-6.7$, $+2.0$]}
\def\SeqCost{\$10.70}
\def\SeqNoClaim{27}
\def\SeqForkCostRatio{1.07}
\def\SeqForkTasksUp{19}
\def\SeqForkTasksTied{44}
\def\SeqForkTasksDown{26}
\def\SeqForkNTrials{445}
\def\SeqForkNGraded{441}
\def\SeqForkNPassed{269}
\def\SeqNGraded{445}
\def\SeqNPassed{282}
\def\SeqForkNLost{4}
\def\SeqForkNTimeouts{38}
\def\SeqForkToolTrials{3}
\def\SeqForkAgentSnapshots{1}
\def\SeqForkRestores{3}
\def\SeqForkRestoresSzero{2}
\def\SeqForkToolPassed{3}
\def\SeqForkCost{\$11.40}

%% file: figures/tab_mcts_numbers.tex
\def\MctsPass{64.0\%}
\def\MctsDeltaSeq{$+0.6$}
\def\MctsDeltaSeqCI{[$-4.5$, $+5.5$]}
\def\MctsDeltaTasks{88}
\def\MctsNGradedTasks{88}
\def\MctsSeqOnCommon{63.2\%}
\def\MctsMixedN{49}
\def\MctsMixedPass{65.4\%}
\def\MctsMixedSeq{64.5\%}
\def\MctsMixedDelta{$+0.9$}
\def\MctsMixedDeltaCI{[$-8.6$, $+10.0$]}
\def\MctsTasksUp{26}
\def\MctsTasksTied{43}
\def\MctsTasksDown{19}
\def\MctsCostRatio{3.81}
\def\MctsTaskMean{63.2\%}
\def\MctsNTrials{444}
\def\MctsNScored{430}
\def\MctsNPassed{275}
\def\MctsNLost{9}
\def\MctsNNoGrade{5}
\def\MctsNLostCkpt{9}
\def\MctsNLostMachine{0}
\def\MctsNTimeouts{0}
\def\MctsCost{\$40.79}
\def\MctsCriticCostShare{31.3\%}
\def\MctsCriticTimeShare{14.0\%}
\def\MctsBranched{99.3\%}
\def\MctsSiblingTies{2.8\%}
\def\MctsSiblingPairs{1684}
\def\MctsUctAgree{90.9\%}
\def\MctsUctPoints{1684}
\def\MctsLaterBranch{70.9\%}
\def\MctsSaturated{68.3\%}
\def\MctsCandidatePass{70.6\%}
\def\MctsNonCandidateN{45}
\def\MctsNonCandidatePass{6.7\%}

%% file: figures/fig_critic_by_position_numbers.tex
\def\CbpNTasks{89}
\def\CbpSamples{5}
\def\CbpNFinal{406}
\def\CbpNoSkill{0.36--0.38}
\def\CbpGapFinal{$-$0.14\,[$-$0.21,\,$-$0.07]}
\def\CbpGapMidMax{0.02}
\def\CbpAcceptAt{50\%}
\def\CbpTrajMaeFinal{39\%}
\def\CbpTrajMaeMidLo{30\%}
\def\CbpTrajMaeMidHi{34\%}
\def\CbpTrajOnFailLo{0.70}
\def\CbpTrajOnFailHi{0.77}
\def\CbpTrajFalseAcceptFinal{5\%}
\def\CbpTrajFalseRejectFinal{60\%}
\def\CbpReadMaeFinal{36\%}
\def\CbpReadMaeMidLo{28\%}
\def\CbpReadMaeMidHi{34\%}
\def\CbpReadOnFailLo{0.47}
\def\CbpReadOnFailHi{0.58}
\def\CbpReadFalseAcceptFinal{10\%}
\def\CbpReadFalseRejectFinal{52\%}
\def\CbpActMaeFinal{22\%}
\def\CbpActMaeMidLo{26\%}
\def\CbpActMaeMidHi{32\%}
\def\CbpActOnFailLo{0.56}
\def\CbpActOnFailHi{0.62}
\def\CbpActFalseAcceptFinal{22\%}
\def\CbpActFalseRejectFinal{22\%}
\def\CbpNMid{1111}
\def\CbpEarlyDepth{25\%}
\def\CbpLateDepth{75\%}
\def\CbpSepMidLo{0.21}
\def\CbpSepMidHi{0.37}
\def\CbpNMidNoLabel{7}
\def\CbpNMidPartial{15}
\def\CbpNFinalNoLabel{4}
\def\CbpNFinalParseFail{1}
\def\CbpTrajChosen{plain}
\def\CbpGapReadTrajEarly{0.00\,[$-$0.05,\,$+$0.06]}
\def\CbpSepGapReadTrajEarly{$+$0.11\,[$+$0.01,\,$+$0.21]}
\def\CbpGapReadTrajMid{0.00\,[$-$0.04,\,$+$0.04]}
\def\CbpSepGapReadTrajMid{$+$0.09\,[$+$0.03,\,$+$0.14]}
\def\CbpGapReadTrajLate{$-$0.01\,[$-$0.04,\,$+$0.01]}
\def\CbpSepGapReadTrajLate{$+$0.09\,[$+$0.04,\,$+$0.13]}
\def\CbpGapReadTrajPooled{0.00\,[$-$0.04,\,$+$0.03]}
\def\CbpSepGapReadTrajPooled{$+$0.10\,[$+$0.05,\,$+$0.15]}
\def\CbpGapReadTrajFinal{$-$0.03\,[$-$0.10,\,$+$0.05]}
\def\CbpSepGapReadTrajFinal{$+$0.03\,[$-$0.10,\,$+$0.15]}
\def\CbpGapReadTrajMidMax{0.01}
\def\CbpGapActTrajEarly{$-$0.02\,[$-$0.05,\,$+$0.00]}
\def\CbpSepGapActTrajEarly{$+$0.11\,[$+$0.04,\,$+$0.18]}
\def\CbpGapActTrajMid{$-$0.01\,[$-$0.03,\,$+$0.02]}
\def\CbpSepGapActTrajMid{$+$0.08\,[$+$0.02,\,$+$0.13]}
\def\CbpGapActTrajLate{$-$0.04\,[$-$0.06,\,$-$0.02]}
\def\CbpSepGapActTrajLate{$+$0.13\,[$+$0.09,\,$+$0.19]}
\def\CbpGapActTrajPooled{$-$0.02\,[$-$0.04,\,$-$0.00]}
\def\CbpSepGapActTrajPooled{$+$0.11\,[$+$0.07,\,$+$0.15]}
\def\CbpGapActTrajFinal{$-$0.17\,[$-$0.24,\,$-$0.09]}
\def\CbpSepGapActTrajFinal{$+$0.20\,[$+$0.08,\,$+$0.31]}
\def\CbpGapActTrajMidMax{0.04}
\def\CbpTrajVarN{3}
\def\CbpTrajVarSpreadMid{1 point}
\def\CbpTrajVarSpreadFinal{3 points}
\def\CbpGapPooled{$-$0.02\,[$-$0.04,\,$+$0.00]}
\def\CbpTrajMaePooled{32\%}
\def\CbpTrajBiasPooled{$+$21}
\def\CbpTrajSepPooled{0.18}
\def\CbpTrajPairPooled{54\%}
\def\CbpTrajFalseAcceptPooled{84\%}
\def\CbpTrajFalseRejectPooled{1\%}
\def\CbpTrajMaeEarly{34\%}
\def\CbpTrajMaeLate{30\%}
\def\CbpTrajPairEarly{49\%}
\def\CbpTrajPairLate{64\%}
\def\CbpTrajBiasMidLo{$+$20}
\def\CbpTrajBiasMidHi{$+$22}
\def\CbpTrajSepFinal{0.36}
\def\CbpReadMaePooled{32\%}
\def\CbpReadBiasPooled{$+$7}
\def\CbpReadSepPooled{0.28}
\def\CbpReadPairPooled{59\%}
\def\CbpReadFalseAcceptPooled{57\%}
\def\CbpReadFalseRejectPooled{14\%}
\def\CbpReadMaeEarly{34\%}
\def\CbpReadMaeLate{28\%}
\def\CbpReadPairEarly{56\%}
\def\CbpReadPairLate{69\%}
\def\CbpReadBiasMidLo{$+$1}
\def\CbpReadBiasMidHi{$+$14}
\def\CbpReadSepFinal{0.38}
\def\CbpActMaePooled{30\%}
\def\CbpActBiasPooled{$+$14}
\def\CbpActSepPooled{0.29}
\def\CbpActPairPooled{59\%}
\def\CbpActFalseAcceptPooled{66\%}
\def\CbpActFalseRejectPooled{5\%}
\def\CbpActMaeEarly{32\%}
\def\CbpActMaeLate{26\%}
\def\CbpActPairEarly{52\%}
\def\CbpActPairLate{68\%}
\def\CbpActBiasMidLo{$+$13}
\def\CbpActBiasMidHi{$+$16}
\def\CbpActSepFinal{0.56}
\def\CbpTrajSepFinalMinusLate{$+$0.11\,[$-$0.01,\,$+$0.24]}
\def\CbpTrajMaeFinalMinusLate{$+$0.09\,[$-$0.02,\,$+$0.21]}
\def\CbpTrajSepFinalMinusPooled{$+$0.17\,[$+$0.05,\,$+$0.29]}
\def\CbpTrajMaeFinalMinusPooled{$+$0.07\,[$-$0.04,\,$+$0.18]}
\def\CbpTrajSkillFinal{$+$0.01\,[$-$0.13,\,$+$0.15]}
\def\CbpTrajSkillMidBest{$-$7}
\def\CbpTrajSepEarly{0.10}
\def\CbpTrajSepLate{0.24}
\def\CbpTrajPairFinal{65\%}
\def\CbpTrajFalseAcceptMidLo{80\%}
\def\CbpTrajFalseAcceptMidHi{91\%}
\def\CbpTrajFalseRejectMidLo{1\%}
\def\CbpTrajFalseRejectMidHi{2\%}
\def\CbpReadSepFinalMinusLate{$+$0.06\,[$-$0.09,\,$+$0.19]}
\def\CbpReadMaeFinalMinusLate{$+$0.08\,[$-$0.02,\,$+$0.18]}
\def\CbpReadSepFinalMinusPooled{$+$0.10\,[$-$0.03,\,$+$0.24]}
\def\CbpReadMaeFinalMinusPooled{$+$0.04\,[$-$0.04,\,$+$0.13]}
\def\CbpReadSkillFinal{$-$0.02\,[$-$0.14,\,$+$0.11]}
\def\CbpReadSkillMidBest{$-$9}
\def\CbpReadSepEarly{0.21}
\def\CbpReadSepLate{0.33}
\def\CbpReadPairFinal{61\%}
\def\CbpReadFalseAcceptMidLo{54\%}
\def\CbpReadFalseAcceptMidHi{61\%}
\def\CbpReadFalseRejectMidLo{4\%}
\def\CbpReadFalseRejectMidHi{28\%}
\def\CbpActSepFinalMinusLate{$+$0.18\,[$+$0.07,\,$+$0.30]}
\def\CbpActMaeFinalMinusLate{$-$0.04\,[$-$0.10,\,$+$0.03]}
\def\CbpActSepFinalMinusPooled{$+$0.27\,[$+$0.16,\,$+$0.38]}
\def\CbpActMaeFinalMinusPooled{$-$0.07\,[$-$0.14,\,$-$0.01]}
\def\CbpActSkillFinal{$-$0.16\,[$-$0.25,\,$-$0.06]}
\def\CbpActSkillMidBest{$-$11}
\def\CbpActSepEarly{0.21}
\def\CbpActSepLate{0.37}
\def\CbpActPairFinal{76\%}
\def\CbpActFalseAcceptMidLo{59\%}
\def\CbpActFalseAcceptMidHi{75\%}
\def\CbpActFalseRejectMidLo{2\%}
\def\CbpActFalseRejectMidHi{8\%}
\def\CbpSkillMidBest{$-$11}
\def\CbpSpreadPooled{0.02}
\def\CbpPairEarlyLo{49\%}
\def\CbpPairEarlyHi{56\%}
\def\CbpPairLateLo{64\%}
\def\CbpPairLateHi{69\%}
\def\CbpFalseAcceptPooledLo{57\%}
\def\CbpFalseAcceptPooledHi{84\%}
\def\CbpFalseRejectPooledHi{14\%}
\def\CbpPairGapReadTrajEarly{$+$6.9 points\,[$-$2.6,\,$+$16.3]}
\def\CbpPairGapReadTrajLate{$+$5.6 points\,[$-$1.8,\,$+$13.1]}
\def\CbpPairGapReadTrajPooled{$+$4.5 points\,[$-$0.7,\,$+$10.0]}
\def\CbpPairGapReadTrajFinal{$-$3.8 points\,[$-$11.9,\,$+$4.7]}
\def\CbpPairDecGapReadTrajPooled{$+$6.8 points\,[$-$2.2,\,$+$15.3]}
\def\CbpSelGapReadTrajPooled{$+$1.4 points\,[$-$2.1,\,$+$5.0]}
\def\CbpPairAccessDiffReadTraj{$-$8.4 points\,[$-$18.0,\,$+$1.3]}
\def\CbpPairGapActTrajEarly{$+$3.2 points\,[$-$5.6,\,$+$11.8]}
\def\CbpPairGapActTrajLate{$+$4.7 points\,[$-$2.7,\,$+$12.1]}
\def\CbpPairGapActTrajPooled{$+$4.4 points\,[$-$0.9,\,$+$9.9]}
\def\CbpPairGapActTrajFinal{$+$11.5 points\,[$+$2.7,\,$+$20.2]}
\def\CbpPairDecGapActTrajPooled{$+$3.4 points\,[$-$5.4,\,$+$12.2]}
\def\CbpSelGapActTrajPooled{$+$1.8 points\,[$-$2.5,\,$+$6.3]}
\def\CbpPairAccessDiffActTraj{$+$7.1 points\,[$-$1.7,\,$+$15.5]}
\def\CbpPairGapActReadEarly{$-$3.8 points\,[$-$12.6,\,$+$4.9]}
\def\CbpPairGapActReadLate{$-$0.9 points\,[$-$10.2,\,$+$7.5]}
\def\CbpPairGapActReadPooled{$-$0.1 points\,[$-$6.0,\,$+$5.8]}
\def\CbpPairGapActReadFinal{$+$15.3 points\,[$+$6.2,\,$+$24.4]}
\def\CbpPairDecGapActReadPooled{$-$3.4 points\,[$-$14.3,\,$+$6.1]}
\def\CbpSelGapActReadPooled{$+$0.4 points\,[$-$3.9,\,$+$4.5]}
\def\CbpPairAccessDiffActRead{$+$15.4 points\,[$+$5.1,\,$+$25.8]}
\def\CbpTrajPairDecPooled{68\%}
\def\CbpTrajSelPooled{4.4}
\def\CbpTrajSelLate{11.0}
\def\CbpReadPairDecPooled{74\%}
\def\CbpReadSelPooled{5.8}
\def\CbpReadSelLate{13.6}
\def\CbpActPairDecPooled{71\%}
\def\CbpActSelPooled{6.2}
\def\CbpActSelLate{13.2}
\def\CbpPairDecPooledLo{68\%}
\def\CbpPairDecPooledHi{74\%}
\def\CbpSelPooledLo{4}
\def\CbpSelPooledHi{6}
\def\CbpSelEarlyLo{$-$2}
\def\CbpSelEarlyHi{$+$3}
\def\CbpSelLateLo{$+$11}
\def\CbpSelLateHi{$+$14}
\def\CbpNPairsPooled{672}
\def\CbpNDecPairsPooled{117}
\def\CbpNPairsFinal{144}
\def\CbpNoiseDecidedShare{79\%}
\def\CbpNoiseSE{10}
\def\CbpNoiseSEMax{22}

%% file: figures/tab_access_class_numbers.tex
\def\AccRecTasks{22}
\def\AccRecN{106}
\def\AccRecTrajFA{5\%}
\def\AccRecTrajFR{46\%}
\def\AccRecTrajFRcount{32}
\def\AccRecTrajFAcount{2}
\def\AccRecTrajNEcount{28}
\def\AccRecTrajNEshare{88\%}
\def\AccRecTokTraj{8}
\def\AccRecReadFA{16\%}
\def\AccRecReadFR{42\%}
\def\AccRecReadFRcount{29}
\def\AccRecReadFAcount{6}
\def\AccRecReadNEcount{27}
\def\AccRecReadNEshare{93\%}
\def\AccRecTokRead{52}
\def\AccRecActFA{19\%}
\def\AccRecActFR{9\%}
\def\AccRecActFRcount{6}
\def\AccRecActFAcount{7}
\def\AccRecActNEcount{3}
\def\AccRecActNEshare{50\%}
\def\AccRecTokAct{64}
\def\AccRecNPass{69}
\def\AccRecNFail{37}
\def\AccRecNEshare{93\%}
\def\AccRecNEcount{27}
\def\AccRecFRcount{29}
\def\AccExeTasks{48}
\def\AccExeN{211}
\def\AccExeTrajFA{3\%}
\def\AccExeTrajFR{64\%}
\def\AccExeTrajFRcount{87}
\def\AccExeTrajFAcount{2}
\def\AccExeTrajNEcount{74}
\def\AccExeTrajNEshare{85\%}
\def\AccExeTokTraj{8}
\def\AccExeReadFA{5\%}
\def\AccExeReadFR{54\%}
\def\AccExeReadFRcount{73}
\def\AccExeReadFAcount{4}
\def\AccExeReadNEcount{53}
\def\AccExeReadNEshare{73\%}
\def\AccExeTokRead{50}
\def\AccExeActFA{19\%}
\def\AccExeActFR{24\%}
\def\AccExeActFRcount{32}
\def\AccExeActFAcount{14}
\def\AccExeActNEcount{8}
\def\AccExeActNEshare{25\%}
\def\AccExeTokAct{64}
\def\AccExeNPass{136}
\def\AccExeNFail{75}
\def\AccExeNEshare{73\%}
\def\AccExeNEcount{53}
\def\AccExeFRcount{73}
\def\AccOraTasks{19}
\def\AccOraN{89}
\def\AccOraTrajFA{7\%}
\def\AccOraTrajFR{68\%}
\def\AccOraTrajFRcount{32}
\def\AccOraTrajFAcount{3}
\def\AccOraTrajNEcount{25}
\def\AccOraTrajNEshare{78\%}
\def\AccOraTokTraj{8}
\def\AccOraReadFA{12\%}
\def\AccOraReadFR{62\%}
\def\AccOraReadFRcount{29}
\def\AccOraReadFAcount{5}
\def\AccOraReadNEcount{17}
\def\AccOraReadNEshare{59\%}
\def\AccOraTokRead{54}
\def\AccOraActFA{31\%}
\def\AccOraActFR{38\%}
\def\AccOraActFRcount{18}
\def\AccOraActFAcount{13}
\def\AccOraActNEcount{5}
\def\AccOraActNEshare{28\%}
\def\AccOraTokAct{79}
\def\AccOraNPass{47}
\def\AccOraNFail{42}
\def\AccOraNEshare{59\%}
\def\AccOraNEcount{17}
\def\AccOraFRcount{29}
\def\AccAllTasks{89}
\def\AccAllN{406}
\def\AccAllTrajFA{5\%}
\def\AccAllTrajFR{60\%}
\def\AccAllTrajFRcount{151}
\def\AccAllTrajFAcount{7}
\def\AccAllTrajNEcount{127}
\def\AccAllTrajNEshare{84\%}
\def\AccAllTokTraj{8}
\def\AccAllReadFA{10\%}
\def\AccAllReadFR{52\%}
\def\AccAllReadFRcount{131}
\def\AccAllReadFAcount{15}
\def\AccAllReadNEcount{97}
\def\AccAllReadNEshare{74\%}
\def\AccAllTokRead{51}
\def\AccAllActFA{22\%}
\def\AccAllActFR{22\%}
\def\AccAllActFRcount{56}
\def\AccAllActFAcount{34}
\def\AccAllActNEcount{16}
\def\AccAllActNEshare{29\%}
\def\AccAllTokAct{67}
\def\AccAllNPass{252}
\def\AccAllNFail{154}
\def\AccAllNEshare{74\%}
\def\AccAllNEcount{97}
\def\AccAllFRcount{131}
\def\AccAllActFRcount{56}
\def\AccActFRwithUnmet{39}
\def\AccActFRsZero{15}
\def\AccReadFRsZero{47}
\def\AccAllActFAcount{34}
\def\AccRecSZeroTasks{21}
\def\AccVarSummaryFA{10\%}
\def\AccVarSummaryFR{54\%}
\def\AccVarChecklistFA{5\%}
\def\AccVarChecklistFR{63\%}
\def\AccVarClaimFA{8\%}
\def\AccVarClaimFR{62\%}

%% file: figures/fig_s0_reference_numbers.tex
\def\SzrNTasks{47}
\def\SzrNStates{212}
\def\SzrNStatesPass{128}
\def\SzrNStatesFail{84}
\def\SzrNProbesPerArm{424}
\def\SzrProbesPerState{2}
\def\SzrAccCtrl{77\%}
\def\SzrAccS{85\%}
\def\SzrAccGap{$+$8.7 points\,[$+$3.3,\,$+$14.5]}
\def\SzrFRCtrl{22\%}
\def\SzrFRS{8\%}
\def\SzrFRGap{$-$14.5 points\,[$-$22.4,\,$-$7.4]}
\def\SzrFACtrl{25\%}
\def\SzrFAS{25\%}
\def\SzrFAGap{0.0 points\,[$-$6.2,\,$+$6.6]}
\def\SzrFRCtrlCount{57}
\def\SzrFRSCount{20}
\def\SzrFACtrlCount{42}
\def\SzrFASCount{42}
\def\SzrFlipCtrl{20\%}
\def\SzrFlipS{7\%}
\def\SzrFlipGap{$-$13.7 points\,[$-$22.2,\,$-$5.6]}
\def\SzrOpenShare{41\%}
\def\SzrStatesHigher{47}
\def\SzrStatesLower{11}
\def\SzrTasksUp{21}
\def\SzrTasksDown{5}
\def\SzrTasksSame{21}
\def\SzrUsedNTasks{28}
\def\SzrUsedNStates{129}
\def\SzrUsedAccCtrl{74\%}
\def\SzrUsedAccS{84\%}
\def\SzrUsedGap{$+$10.9 points\,[$+$2.6,\,$+$19.6]}
\def\SzrUsedOpenShare{64\%}
\def\SzrUsedFlipCtrl{26\%}
\def\SzrUsedFlipS{9\%}
\def\SzrUnusedNTasks{19}
\def\SzrUnusedNStates{83}
\def\SzrUnusedAccCtrl{81\%}
\def\SzrUnusedAccS{87\%}
\def\SzrUnusedGap{$+$5.4 points\,[$+$1.8,\,$+$9.1]}
\def\SzrUnusedOpenShare{4\%}
\def\SzrUnusedFlipCtrl{11\%}
\def\SzrUnusedFlipS{2\%}
\def\SzrGapCtrlFirst{$+$4.9 points\,[$+$0.6,\,$+$9.2]}
\def\SzrGapSFirst{$+$2.7 points\,[$-$1.5,\,$+$6.9]}
\def\SzrOtherNTasks{42}
\def\SzrOtherNStates{192}
\def\SzrOtherAccCtrl{75\%}
\def\SzrOtherAccS{74\%}
\def\SzrOtherAccGap{$-$1.3 points\,[$-$4.4,\,$+$1.8]}
\def\SzrOtherFRCtrl{26\%}
\def\SzrOtherFRS{27\%}
\def\SzrOtherFACtrl{24\%}
\def\SzrOtherFAS{24\%}
\def\SzrOtherFlipCtrl{16\%}
\def\SzrOtherFlipS{11\%}
\def\SzrOtherOpenShare{7\%}
\def\SzrOtherTasksUp{8}
\def\SzrOtherTasksDown{9}
\def\SzrOtherTasksSame{25}
\def\SzrAllNTasks{89}
\def\SzrAllNStates{404}
\def\SzrAllAccCtrl{76\%}
\def\SzrAllAccS{80\%}
\def\SzrAllAccGap{$+$4.0 points\,[$+$0.6,\,$+$7.5]}
\def\SzrAllFRCtrl{24\%}
\def\SzrAllFRS{17\%}
\def\SzrAllFRGap{$-$6.5 points\,[$-$11.3,\,$-$2.2]}
\def\SzrAllFACtrl{24\%}
\def\SzrAllFAS{25\%}
\def\SzrAllFAGap{$+$0.3 points\,[$-$4.4,\,$+$5.0]}
\def\SzrAllFlipCtrl{18\%}
\def\SzrAllFlipS{9\%}
\def\SzrAllFlipGap{$-$9.4 points\,[$-$15.4,\,$-$3.8]}
\def\SzrAllOpenShare{25\%}
\def\SzrAllTasksUp{29}
\def\SzrAllTasksDown{14}
\def\SzrAllTasksSame{46}
\def\SzrForkNStates{121}
\def\SzrForkNTasks{37}
\def\SzrForkNEvals{199}
\def\SzrForkAccCtrl{71\%}
\def\SzrForkAccS{81\%}
\def\SzrForkAccGap{$+$9.9 points\,[$+$2.0,\,$+$18.2]}
\def\SzrForkFRCtrl{28\%}
\def\SzrForkFRS{12\%}
\def\SzrForkFRGap{$-$15.6 points\,[$-$26.0,\,$-$4.2]}
\def\SzrForkFACtrl{29\%}
\def\SzrForkFAS{30\%}
\def\SzrForkFlipCtrl{26\%}
\def\SzrForkFlipS{11\%}
\def\SzrForkStateShare{30\%}
\def\SzrNoForkNStates{283}
\def\SzrNoForkAccCtrl{78\%}
\def\SzrNoForkAccS{79\%}
\def\SzrNoForkAccGap{$+$1.4 points\,[$-$1.3,\,$+$4.1]}
\def\SzrNoForkFlipCtrl{15\%}
\def\SzrNoForkFlipS{8\%}
\def\SzrNRecorded{411}
\def\SzrNKept{404}
\def\SzrNDropped{7}
\def\SzrDropUnfinished{3}
\def\SzrDropNoLabel{1}
\def\SzrDropNoFork{3}
\def\SzrNReplayChecked{401}
\def\SzrReplayMismatch{2}
\def\SzrReplayMatch{399}
\def\SzrNNoRecorded{3}
\def\SzrRcMismatchStates{10}

%% file: figures/fig_critic_by_position.tex
\input{figures/fig_critic_by_position_numbers}
\begin{figure}[!htb]
\centering
\includegraphics[width=\linewidth]{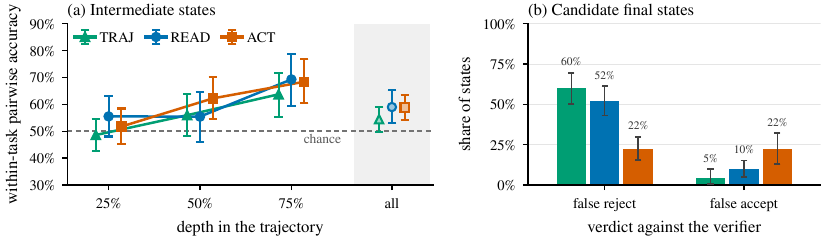}
\caption{Critics by position in the actor's trajectory, on \CbpNTasks{} Terminal-Bench~2.1 tasks (numbers in \Cref{app:critic-by-position}); \textsc{traj}, \textsc{read}, \textsc{act} are $C^{\mathrm{traj}}$, $C^{\mathrm{read}}$, $C^{\mathrm{act}}$ (\S\ref{sec:q-act}), \textsc{traj} with the plain prompt (variants in \Cref{app:traj-variants}). \textbf{(a)}~Share of same-task, same-depth pairs that $\hat q$ orders like the pass rates of their \CbpSamples{} continuations; \emph{all} pools the depths; the dashed line marks chance (50\%). \textbf{(b)}~Passing states that $\hat v$ rejects and failing states it accepts.}
\label{fig:critic-by-position}
\end{figure}

%% file: figures/tab_access_class_compact.tex
\input{figures/tab_access_class_numbers}
\begin{table}[!ht]  %
\caption{Verdicts at candidate final states against the verifier, over \AccAllN{} states of \AccAllTasks{} tasks, grouped by what the verifier does to decide a task: it \emph{reads} files and compares them with a value or format it stores (\AccRecTasks{} tasks), \emph{runs} the artifact or probes a service (\AccExeTasks), or \emph{holds the answer}, a value that cannot be recovered from the environment (\AccOraTasks). FR: share of passing states rejected. FA: share of failing states accepted. Model cost per state in \Cref{app:critic-cost}.}
\label{tab:access-class}
\vspace{4pt}
\centering
\small
\setlength{\tabcolsep}{4.5pt}
\begin{tabular}{@{}lcccccccc@{}}
 & \multicolumn{2}{c}{reads} & \multicolumn{2}{c}{runs} & \multicolumn{2}{c}{holds the answer} & \multicolumn{2}{c}{all} \\
\cmidrule(lr){2-3}\cmidrule(lr){4-5}\cmidrule(lr){6-7}\cmidrule(l){8-9}
 & FR$\downarrow$ & FA$\downarrow$ & FR$\downarrow$ & FA$\downarrow$ & FR$\downarrow$ & FA$\downarrow$ & FR$\downarrow$ & FA$\downarrow$ \\
\midrule
$C^{\mathrm{traj}}$ & \AccRecTrajFR & \AccRecTrajFA & \AccExeTrajFR & \AccExeTrajFA & \AccOraTrajFR & \AccOraTrajFA & \AccAllTrajFR & \AccAllTrajFA \\
$C^{\mathrm{read}}$ & \AccRecReadFR & \AccRecReadFA & \AccExeReadFR & \AccExeReadFA & \AccOraReadFR & \AccOraReadFA & \AccAllReadFR & \AccAllReadFA \\
$C^{\mathrm{act}}$ & \AccRecActFR & \AccRecActFA & \AccExeActFR & \AccExeActFA & \AccOraActFR & \AccOraActFA & \AccAllActFR & \AccAllActFA \\
\end{tabular}
\end{table}

%% file: figures/fig_s0_reference_bars.tex
\input{figures/fig_s0_reference_numbers}
\begin{figure}[!ht]
\centering
\begin{minipage}[t]{0.485\linewidth}
\centering
\includegraphics[width=\linewidth]{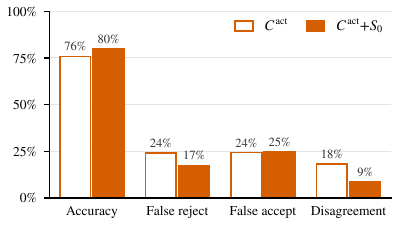}
\caption{$C^{\mathrm{act}}$ against $C^{\mathrm{act}}{+}\Sinit$ on all \SzrAllNStates{} candidate final states. Each critic judged every state twice, and every rate counts both verdicts. Disagreement is the share of states on which one critic's own two verdicts differ.}
\label{fig:s0-reference-all}
\end{minipage}\hfill
\begin{minipage}[t]{0.485\linewidth}
\centering
\includegraphics[width=\linewidth]{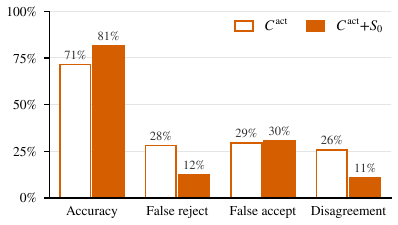}
\caption{The same comparison on the \SzrForkNStates{} states on which $C^{\mathrm{act}}{+}\Sinit$ inspected the fork of the initial state in at least one of its two evaluations.}
\label{fig:s0-reference}
\end{minipage}
\end{figure}

%% file: sections/04_acting_critic.tex
\section{\texorpdfstring{\sys}{Probe-and-Resume}}
\label{sec:acting}

\input{figures/fig_acting_loop_numbers}
The previous section identified a number of unique design decisions for critics operating in branchable terminal environments.  We found that their value estimates for intermediate states are no better than chance early in a trajectory and improve little with execution access, while unrestricted execution in a fork of the current state and of the initial state helps them judge candidate final states.  
Based on these findings, we introduce the \sys{} test-time search policy for terminal agents, and report on its end-to-end evaluations on terminal benchmarks.

\input{figures/fig_acting_loop_diagram}

\subsection{How \sys Works}
\label{sec:acting-design}

Figure~\ref{fig:loop} shows the search policy.   
The actor starts in $\Sinit$ and executes as normal without knowledge of the critic.   When it decides to call \texttt{task\_complete} at $S^{(1)}$, we use $C^{\mathrm{act}}{+}\Sinit$ (\S\ref{sec:q-info}) to evaluate the proposal. 
The critic lists the requirements in the task and executes in the disposable forks of $S^{(1)}$ and $\Sinit$ to check each requirement.  
Thus, the critic can delete files, regenerate files, copy data between forks, install interpreters, start or kill processes, and take other potentially destructive actions.

If any requirements fail, the actor receives the failures, the rationale for why they failed, and how the critic checked.  
The actor resumes from $S^{(1)}$, and observes from the \texttt{task\_complete} call that the state was rejected, the critic's feedback, and a request to either address the issues or submit a command that proves a failed requirement actually holds.
The actor works until it calls \texttt{task\_complete} again and then submits another proposal.    We write $S^{(r)}$ for the $r^{\text{th}}$ proposed state. 

When the actor submits $S^{(2)}$, the critic continues its conversation from when it assessed $S^{(1)}$. It is told that the actor has revised its work, is shown the commands the actor ran, and is given fresh forks.  It resumes from its previous requirements list and checks the entire list again.  
If the critic accepts the state, it is submitted to the verifier as the final result.

We execute search until the critic accepts a state, until six critic rounds have passed, or until a budget runs out.  The task has a total time limit of $T$, shared by the actor's turns, the critic's evaluations and every snapshot, restore and fork; only the verifier's grading is not charged. After the first rejection, the actor also has at most 50 more turns.
If the critic did not accept any state, the last proposed state is submitted to the verifier; if no such state exists, it is considered a failure.

\subsection{Terminal-Bench 2.1}
\label{sec:acting-results}

We run all \LoopNTasks{} Terminal-Bench~2.1 tasks 5 times each, using GPT-5.6 Luna as the model in all settings.  We compare with a baseline that submits the first proposed final state, the Codex CLI harness, and \sys.      Codex results are drawn from Harbor Hub's public leaderboard: Codex CLI with GPT-5.6 Luna on the same 89 tasks, five trials each, under the same per-task time limits. Two differences favor Codex. Its $T$ covers only the agent, while ours also covers the critic, and a Codex trial that reaches $T$ is graded on the state it left, while ours fails unless it has proposed a state. It also uses its own harness and sandboxes and calls the model through OpenAI's API rather than Bedrock. We count one Codex trial whose verifier timed out as a failure, and price its tokens at our rate.

Table~\ref{tab:acting-results} shows that \sys improves by 9 percentage points ($\LoopRoundZero\to \LoopHandIn$) over the baseline, and is close to Codex CLI's performance.   It improves the pass rate on 30/89 tasks and regresses on 1.  The improvements are comparable when using Claude Sonnet 5 (\LoopXSonnetGainHandInPts{} points) or GLM-5 (\LoopXGlmGainHandInPts{} points) as the critic's model (\Cref{app:crossmodel}).  
While \sys increases the cost to $2.2\times$ the baseline's ($\$10.70\to \$23.14$), it remains half the cost of Codex and only \LoopBelowCodexPts{} points lower in accuracy.   This is because the critic is only called when the actor believes it is done, rather than during many intermediate turns. To separate resuming from the extra round of feedback, we also restart the actor from $\Sinit$ after each rejection, with the same critic, feedback and time limit (\Cref{app:restart}). Restarting costs \RstCostRatio$\times$ as much as resuming and takes longer, and its pass rate is not higher (Table~\ref{tab:acting-results}).

\input{figures/tab_acting_results}

\paragraph{Detailed Analysis.}
\label{sec:acting-mechanism}

To understand why the critic improves the pass rate,  Figure~\ref{fig:acting-loop}a depicts a flow chart between the true pass rate of the first state the actor proposes to the critic, the critic's assessment of it, and the true pass rate of the final submitted state.  Of the \LoopFirstFail{} proposals that fail, the critic correctly rejects \LoopFailRejected{} (\LoopCatchRate), and the actor eventually is able to fix and ``rescue'' \LoopRescued{} of them (\LoopRescueOfCaught). 
Of the \LoopFirstPass{} passing proposals, the critic falsely rejects \LoopPassRejected{} (\LoopFirstFR) and the actor eventually ``rescues'' \LoopFRSurvived{}. 

\input{figures/fig_acting_loop}

Of the \LoopFailSubmitted{} submissions that fail, \LoopFailAccepted{} were the critic's false accepts, the critic timed out on \LoopFailNoVerdict, and \LoopFRLost{} were falsely rejected.  The other \LoopRejectedNotRescued{} were correctly rejected by the critic, but the actor was not able to improve them. The remaining \LoopNeverClaimed{} ran out of time before the actor's first claim, so the critic never saw them.
Of the trials that claim to be finished, \LoopOneRoundShare{} end after one round. Allowing a second claim raises the submitted pass rate from \LoopByRoundOnePct{} to \LoopByRoundTwoPct, \LoopGainByRoundTwoShare{} of the loop's gain, and a third to \LoopByRoundThreePct; further rounds add nothing measurable (Figure~\ref{fig:acting-loop}b), so a cap of three claims would lose nothing.

The critic's false accept and reject rates (21\%, 21\%) are comparable to those measured in \S\ref{sec:q-act} for $C^{\mathrm{act}}$ at candidate final states (\LoopOfflineActFA{}, \LoopOfflineActFR).  A benefit of \sys is that after a false rejection the actor does not need to start over from the initial state, and can resume immediately.

\subsection{SWE-bench Pro}
\label{app:swepro}

To study whether \sys generalizes to non-Terminal-Bench tasks, we then evaluate \sys{} (unchanged) with GPT-5.6 Luna on SWE-bench Pro~\citep{deng2025swebenchpro}, using the public split of its second release, V2.0.0 (642 tasks from 11 open-source repositories).
We use two task sets, both defined by data released with the benchmark: \emph{Hard}, its HARD-51 subset, and \emph{Medium}, tasks that none of its runs of 2025 models solved. We report one trial of each of the \SweProNTasks{} tasks we can grade offline (\SweProNHardScored{} hard, \SweProNMediumScored{} medium) under the 50-minute limit; \Cref{app:swepro-setup} gives the exact sets and four more trials of each hard task.

V2's locked protocol runs the agent offline except for the model endpoint, as we do, but grades the agent's \texttt{git diff} applied to a fresh image. We instead run the same, unchanged verifier on a fork of the proposed state, as on Terminal-Bench. The verifier first resets the repository's test code. The hidden tests enter only that fork, and neither the actor nor the critic sees them or the grade. Since the sandbox had no network, no trial could download packages.

Figure~\ref{fig:swepro} shows that while the baseline that submits the first proposed state has a pass rate of \SweProFirstPct, \sys increases the pass rate to \SweProSubmittedPct, comparable to the improvements on Terminal-Bench; the gain comes from the medium tasks, and on the hard tasks it cannot be told apart from zero (\Cref{app:swepro-setup}).   
The benefits are similar because the critic rejects \SweProCaught{} of the \SweProFailFirst{} failing first proposals (\SweProCatchRate) and the actor turns \SweProRescued{} of them into passing states. The critic also falsely rejects \SweProFalseReject{} of the \SweProPassFirst{} passing first proposals  (\SweProFalseRejectRate), and the actor is able to rescue \SweProFalseRejectKept{} of them. 
It costs \SweProCostRatio$\times$ as much.

\input{figures/fig_swepro}

\subsection{Discussion and Limitations}

\paragraph{Branching for assessment and repair.}
This work shows that branching is not only useful for exploring alternative actions, but also for critics to evaluate a candidate without changing the state the actor may resume from.  It also allows the critic to compare with other states as a reference (such as the initial state).   We find that critics need not be fully correct, as the actor and critic pair can ``rescue'' seemingly failing states.  The actor also avoids restarting from the beginning after a critic rejection, which lowers success by \RstLossHandInPts{} points and increases model cost (\Cref{app:restart}).   %

\paragraph{Scope and model dependence.}
Critics help when actors can repair mistakes they fail to identify themselves; easy tasks the actor can solve leave little room for critics to help, while tasks that are too hard may not be solvable by the actor.  As weak evidence, we find no statistically significant gains using Astra and Opus as actor or critic on even hard Terminal-Bench tasks (\Cref{app:stronger}). Our evaluation manages environment state using Waypoint's terminal-oriented container. Other settings require branching support for their state, such as TClone for GUI workspaces~\citep{huang2026tclone}. Forks do not isolate writes to external databases or shared services, so destructive checks require isolating those dependencies or restricting critic actions.

\paragraph{Sequential and parallel search.}
Probe-and-Resume spends test-time compute sequentially.
Parallel search spends compute on breadth instead, drawing $k$ independent trials and selecting one of them.
The two are complementary as each of the $k$ trials can itself be a Probe-and-Resume loop, and the acting critic's final verdicts can evaluate on forks of each trial's environment state to select a winner.
We found that pass rates converge after three rounds (\S\ref{sec:acting-mechanism}), so compute may be better spent on parallel trials.  We leave studying this interaction to future work.

\paragraph{Towards Complex Computer Tasks.}
As computer tasks involve richer data and interacting systems, correctness will increasingly encompass the entire environment state {\it and} how it changes. Existing benchmarks already point towards this: AppWorld checks for unintended database changes~\citep{trivedi2024appworld}, ToolSandbox evaluates dependencies and required events throughout execution~\citep{lu2025toolsandbox}, and STRATUS checks whether interventions worsen system health and relies on undoing them~\citep{chen2025stratus}. %
Even as models increase in capability, they will likely benefit from branchable environments, as well as the critic's ability to effectively use branching capabilities to execute checks within and across intermediate and final environment states.

%% file: figures/fig_acting_loop_diagram.tex
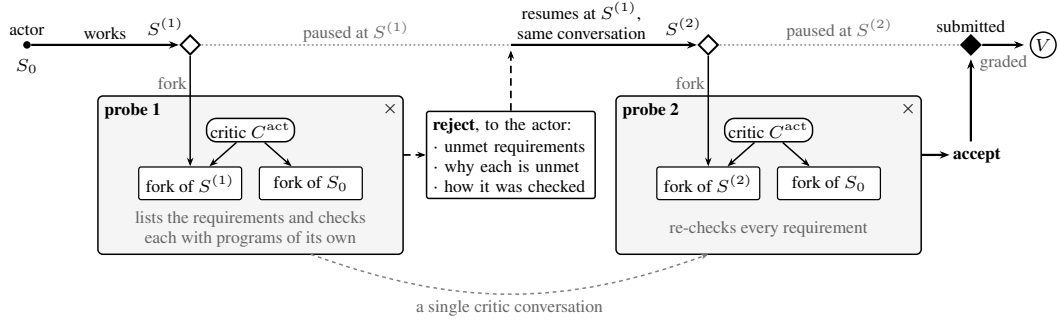
\begin{figure}[!htb]
\centering
\resizebox{\linewidth}{!}{%
\begin{tikzpicture}[
  x=1cm,y=1cm,
  ex/.style     = {line width=0.9pt, black, -{Stealth[length=4pt]}},
  wait/.style   = {line width=0.7pt, black!40, densely dotted},
  fk/.style     = {line width=0.6pt, black, -{Stealth[length=3.5pt]}},
  fb/.style     = {line width=0.7pt, black, densely dashed, -{Stealth[length=3.5pt]}},
  mem/.style    = {line width=0.6pt, black!50, dash pattern=on 1.5pt off 1.5pt, -{Stealth[length=3pt]}},
  st/.style     = {circle, fill=black, inner sep=0pt, minimum size=3pt},
  dc/.style     = {diamond, draw=black, fill=white, line width=0.8pt, inner sep=0pt, minimum size=8pt},
  dcg/.style    = {diamond, draw=black, fill=black, line width=0.8pt, inner sep=0pt, minimum size=8pt},
  probe/.style  = {draw=black, rounded corners=2pt, line width=0.6pt, fill=black!4},
  mach/.style   = {draw=black, fill=white, rounded corners=1pt, line width=0.5pt, minimum width=1.45cm,
                   minimum height=0.46cm, font=\scriptsize, align=center, inner sep=1pt},
  crit/.style   = {draw=black, fill=white, rounded corners=4pt, line width=0.6pt, inner sep=1.6pt, font=\scriptsize},
  vcall/.style   = {draw=black, circle, line width=0.6pt, inner sep=0.8pt, font=\scriptsize},
  lb/.style     = {font=\scriptsize, align=center, inner sep=1pt},
  sb/.style     = {font=\scriptsize, align=center, inner sep=1pt, black!60},
  tg/.style     = {font=\scriptsize\bfseries, inner sep=1pt},
  msg/.style    = {draw=black, fill=white, line width=0.6pt, rounded corners=1pt, font=\scriptsize, align=left,
                   inner sep=2.5pt, text width=2.2cm},
]

\node[st] (s0) at (0,0) {};
\node[lb] at (0,-0.3) {$S_0$};
\node[lb, anchor=south] at (0,0.1) {actor};
\draw[ex] (s0) -- node[lb, above=1pt] {works} (2.14,0);
\node[dc] (c1) at (2.3,0) {};
\node[lb, anchor=south east] at (2.25,0.1) {$S^{(1)}$};
\node[dc] (c2) at (9.6,0) {};
\node[lb, anchor=south east] at (9.55,0.1) {$S^{(2)}$};
\draw[wait] (9.76,0) -- (13.14,0);
\node[sb, anchor=south] at (11.45,0.03) {paused at $S^{(2)}$};
\node[dcg] (sub) at (13.3,0) {};
\node[lb, anchor=south] at (13.3,0.14) {submitted};
\draw[ex] (13.46,0) -- (14.05,0);
\node[vcall] (v) at (14.32,0) {$V$};
\node[sb, anchor=north] at (13.76,-0.1) {graded};

\draw[probe] (1.0,-0.72) rectangle (5.3,-2.95);
\node[tg, anchor=north west] at (1.05,-0.77) {probe 1};
\node[crit] (k1) at (3.15,-1.2) {critic $C^{\mathrm{act}}$};
\node[mach] (m1) at (2.3,-1.95) {fork of $S^{(1)}$};
\node[mach] (o1) at (4.0,-1.95) {fork of $S_0$};
\draw[fk] (k1) -- (m1);
\draw[fk] (k1) -- (o1);
\draw[fk] (c1) -- node[sb, left=0pt, pos=0.25] {fork} (m1);
\node[sb] at (3.15,-2.45) {lists the requirements and checks};
\node[sb] at (3.15,-2.74) {each with programs of its own};
\node[lb] at (5.12,-0.9) {$\times$};

\node[msg, anchor=west] (fbk) at (5.62,-1.55) {\textbf{reject}, to the actor:\\[1pt]
  $\cdot$ unmet requirements\\ $\cdot$ why each is unmet\\ $\cdot$ how it was checked};
\draw[fb] (5.3,-1.55) -- (fbk.west);
\coordinate (r1) at (fbk.north |- 0,0);
\draw[fb] (fbk.north) -- ([yshift=-2.2pt]r1);
\draw[wait] (2.46,0) -- (r1);
\node[sb, anchor=south] at ($(2.46,0)!0.5!(r1)$) {paused at $S^{(1)}$};
\draw[ex] (r1) -- node[lb, above=1pt, pos=0.38] {resumes at $S^{(1)}$,\\ same conversation} (9.44,0);

\draw[probe] (8.3,-0.72) rectangle (12.6,-2.95);
\node[tg, anchor=north west] at (8.35,-0.77) {probe 2};
\node[crit] (k2) at (10.45,-1.2) {critic $C^{\mathrm{act}}$};
\node[mach] (m2) at (9.6,-1.95) {fork of $S^{(2)}$};
\node[mach] (o2) at (11.3,-1.95) {fork of $S_0$};
\draw[fk] (k2) -- (m2);
\draw[fk] (k2) -- (o2);
\draw[fk] (c2) -- node[sb, left=0pt, pos=0.25] {fork} (m2);
\node[sb] at (10.45,-2.6) {re-checks every requirement};
\node[lb] at (12.42,-0.9) {$\times$};
\draw[ex] (12.6,-1.55) -- (13.0,-1.55) node[tg, anchor=west] {accept};
\draw[ex] (13.3,-1.36) -- (13.3,-0.2);

\draw[mem] (4.0,-2.95) to[out=-20,in=200] node[sb, below=1pt] {a single critic conversation} (9.6,-2.95);

\end{tikzpicture}
}
\caption{\sys. When the actor claims to be done, the critic uses forks of $S^{(r)}$ and $S_0$ and can freely execute actions to evaluate the state; they are then discarded ($\times$ symbol).  If the critic rejects, the actor resumes at $S^{(r)}$ with the critic's feedback.  Otherwise, $S^{(r)}$ goes to the verifier $V$.}
\label{fig:loop}
\end{figure}

%% file: figures/tab_acting_results.tex
\begin{table}[!htb]
\caption{Pass@1 on Terminal-Bench~2.1 over 5 trials per task; cost is the model cost of one trial of every task at the AWS Bedrock rate (\Cref{app:prices}). Restart from $\Sinit$ restarts the actor from $\Sinit$ after each rejection (\Cref{app:restart}); Codex CLI is Harbor Hub's job (\S\ref{sec:acting-results}), priced at the same rate.}
\label{tab:acting-results}
\vspace{4pt}
\centering
\small
\input{figures/tab_acting_results_body}
\end{table}

%% file: figures/tab_acting_results_body.tex
\begin{tabular}{@{}lccc@{}}
 & \textbf{Pass@1} & \textbf{$\Delta$ vs.\ baseline [95\% CI]} & \textbf{Total model cost} \\
Baseline & 63.4\% & -- & \$10.70 \\
\sys & \textbf{72.4\%} & $+$9.0\,[$+$5.6,\,$+$12.6] & \$23.14 \\
Restart from $\Sinit$ & 66.3\% & -- & \$38.12 \\
\midrule
Codex CLI (reference) & 76.6\% & -- & \$50.29 \\
\end{tabular}

%% file: figures/fig_acting_loop.tex
\begin{figure}[!ht]  %
\centering
\includegraphics[width=\linewidth]{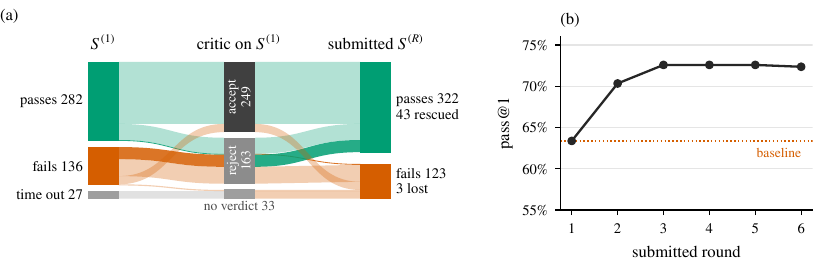}
\caption{(a) Flow diagram that describes the true pass rate of the first state the actor proposes to the critic  $S^{(1)}$, the critic's assessment of it, and the true pass rate of the final state submitted to the verifier.  Trials that exceed the time limit count as failures.  (b) Pass@1 if stopped after $K$ rounds.  }
\label{fig:acting-loop}
\end{figure}

%% file: figures/fig_swepro.tex
\begin{figure}[!ht]  %
\centering
\begin{minipage}[c]{0.64\linewidth}
\includegraphics[width=\linewidth]{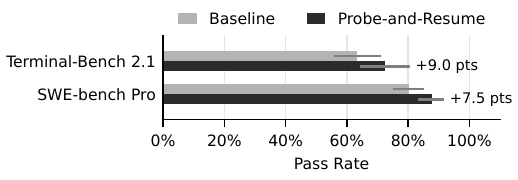}
\end{minipage}\hfill
\begin{minipage}[c]{0.33\linewidth}
\caption{\sys{} on Terminal-Bench 2.1 and Medium/Hard tasks in SWE-bench Pro.}
\label{fig:swepro}
\end{minipage}
\end{figure}

%% file: sections/05_conclusion.tex
\section{Conclusion}
We studied when terminal-agent critics provide useful assessments and what access they need. The tested critics make weak predictions of success from intermediate states, even with execution access. Once the actor claims to be done, a critic that runs checks in a fork of that state judges it far more accurately. A fork of the initial state also lets it check what must be preserved or recovered. These findings motivate \sys: the critic checks candidate completions in disposable forks, and the actor resumes from the unchanged candidate with feedback after rejection. On Terminal-Bench~2.1, \sys{} raises pass@1 from $63.4\%$ to $72.4\%$, at less than half the cost of Codex CLI with the same model. These results show that branching alone does not make search effective; it helps when a critic uses it to check completed work, because evaluating a terminal task is itself a terminal task.

%% file: appendix/A_prompts.tex
\section{Prompts}
\label{app:prompts}

All prompts used in this paper are in our repository at \coderepo.
The actor's prompts are under \texttt{src/harbor/search/templates/}, whereas the critics' prompts and the feedback template are under \texttt{src/harbor/search/critics/templates/}.
The code in the same directories shows how each prompt is built from its template, the task and the trajectory.

%% file: appendix/B_tasks.tex
\section{Task List, Verifier Annotation and Timeouts}
\label{app:tasks}

Table~\ref{tab:tasks} lists the 89 tasks of Terminal-Bench~2.1, the time limit $T$ each task gives the actor, and what each task's verifier does to grade it.

We read every verifier by hand. For each assertion it makes, we asked what a checker must do, at minimum, to settle it: read something that already exists in the environment (R), run something to produce a new observation (X), or compare against a value that only the verifier holds (O). R and X have finer codes. R1: file content, such as a file that exists or whose content or hash matches. R2: filesystem metadata, such as permissions, owners, timestamps or git objects. R3: runtime state, such as a live process, a listening port or an environment variable. X1: run a binary, a test or a computation and check its output. X2: probe a service, such as connecting to a port or sending a request. A task carries every code its assertions need.

\S\ref{sec:q-act} groups tasks by their strongest code: \emph{holds the answer} if any assertion is O (\AccOraTasks{} tasks), otherwise \emph{runs} if any is X (\AccExeTasks), otherwise \emph{reads} (\AccRecTasks). The last column marks the \SzrNTasks{} tasks whose instruction states a requirement relative to the initial state $\Sinit$, such as a file that must stay unchanged or an output that must match the original program's.

\begin{table}[!htb]
\caption{The 89 tasks of Terminal-Bench~2.1. $T$: the task's time limit for the actor. \emph{Verifier}: every code its assertions need (R1--R3 read, X1--X2 run, O compares with a value only the verifier holds); \S\ref{sec:q-act} groups tasks by the strongest one. $\Sinit$: the task states a requirement relative to the initial state (\S\ref{sec:q-info}).}
\label{tab:tasks}
\vspace{4pt}
\centering
\scriptsize
\input{figures/tab_tasks_body}
\end{table}

%% file: figures/tab_tasks_body.tex
\setlength{\tabcolsep}{3pt}
\resizebox{\textwidth}{!}{%
\begin{tabular}{@{}l r l c@{}}
\toprule
Task & $T$ (min) & Verifier & $\Sinit$ \\
\midrule
\texttt{adaptive-rejection-sampler} & 15 & R1, X1 &  \\
\texttt{bn-fit-modify} & 60 & R1, O &  \\
\texttt{break-filter-js-from-html} & 20 & R1, X1 &  \\
\texttt{build-cython-ext} & 15 & R1, X1, O & \checkmark \\
\texttt{build-pmars} & 15 & R1, X1 &  \\
\texttt{build-pov-ray} & 200 & R1, X1, O & \checkmark \\
\texttt{caffe-cifar-10} & 60 & R1, R2, X1 &  \\
\texttt{cancel-async-tasks} & 15 & R1, X1 &  \\
\texttt{chess-best-move} & 15 & R1 & \checkmark \\
\texttt{circuit-fibsqrt} & 60 & R1, X1 & \checkmark \\
\texttt{cobol-modernization} & 15 & R1, X1 & \checkmark \\
\texttt{code-from-image} & 20 & R1 & \checkmark \\
\texttt{compile-compcert} & 40 & R1, R2, X1 &  \\
\texttt{configure-git-webserver} & 15 & X1, X2 &  \\
\texttt{constraints-scheduling} & 20 & R1 & \checkmark \\
\texttt{count-dataset-tokens} & 15 & R1, O &  \\
\texttt{crack-7z-hash} & 30 & R1 & \checkmark \\
\texttt{custom-memory-heap-crash} & 30 & R1, X1 & \checkmark \\
\texttt{db-wal-recovery} & 15 & R1 & \checkmark \\
\texttt{distribution-search} & 60 & R1 &  \\
\texttt{dna-assembly} & 30 & R1, X1 & \checkmark \\
\texttt{dna-insert} & 30 & R1, X1 & \checkmark \\
\texttt{extract-elf} & 15 & R1, X1, O &  \\
\texttt{extract-moves-from-video} & 30 & R1 & \checkmark \\
\texttt{feal-differential-cryptanalysis} & 30 & X1, O &  \\
\texttt{feal-linear-cryptanalysis} & 30 & R1 & \checkmark \\
\texttt{filter-js-from-html} & 30 & R1, X1, X2, O &  \\
\texttt{financial-document-processor} & 20 & R1 & \checkmark \\
\texttt{fix-code-vulnerability} & 15 & R1, X1 & \checkmark \\
\texttt{fix-git} & 15 & R1 & \checkmark \\
\texttt{fix-ocaml-gc} & 60 & R1, X1, X2, O &  \\
\texttt{gcode-to-text} & 15 & R1 & \checkmark \\
\texttt{git-leak-recovery} & 15 & R1, R2 & \checkmark \\
\texttt{git-multibranch} & 15 & X2 &  \\
\texttt{gpt2-codegolf} & 15 & R2, X1 & \checkmark \\
\texttt{headless-terminal} & 15 & R1, X1, X2 &  \\
\texttt{hf-model-inference} & 15 & R1, X1, X2 &  \\
\texttt{install-windows-3.11} & 60 & R1, R3, X2 & \checkmark \\
\texttt{kv-store-grpc} & 15 & R1, R3, X2 &  \\
\texttt{large-scale-text-editing} & 20 & R1, X1 & \checkmark \\
\texttt{largest-eigenval} & 15 & X1 &  \\
\texttt{llm-inference-batching-scheduler} & 30 & R1 & \checkmark \\
\texttt{log-summary-date-ranges} & 15 & R1 & \checkmark \\
\texttt{mailman} & 30 & R1, X2 &  \\
\texttt{make-doom-for-mips} & 15 & R1, X1 & \checkmark \\
\bottomrule
\end{tabular}\qquad
\begin{tabular}{@{}l r l c@{}}
\toprule
Task & $T$ (min) & Verifier & $\Sinit$ \\
\midrule
\texttt{make-mips-interpreter} & 30 & R1, X1 & \checkmark \\
\texttt{mcmc-sampling-stan} & 30 & R1, X1 & \checkmark \\
\texttt{merge-diff-arc-agi-task} & 15 & R1, X1, O & \checkmark \\
\texttt{model-extraction-relu-logits} & 15 & R1, X1 &  \\
\texttt{modernize-scientific-stack} & 10 & X1 & \checkmark \\
\texttt{mteb-leaderboard} & 60 & R1, O &  \\
\texttt{mteb-retrieve} & 30 & R1 & \checkmark \\
\texttt{multi-source-data-merger} & 15 & R1 & \checkmark \\
\texttt{nginx-request-logging} & 15 & R1, X1, X2 &  \\
\texttt{openssl-selfsigned-cert} & 15 & R1, R2, X1 &  \\
\texttt{overfull-hbox} & 12.5 & R1, X1 & \checkmark \\
\texttt{password-recovery} & 15 & R1 & \checkmark \\
\texttt{path-tracing} & 30 & R1, X1 & \checkmark \\
\texttt{path-tracing-reverse} & 30 & R1, X1 & \checkmark \\
\texttt{polyglot-c-py} & 15 & X1 &  \\
\texttt{polyglot-rust-c} & 15 & R1, X1 &  \\
\texttt{portfolio-optimization} & 60 & R1, X1 & \checkmark \\
\texttt{protein-assembly} & 30 & R1, O & \checkmark \\
\texttt{prove-plus-comm} & 15 & R1, X1 &  \\
\texttt{pypi-server} & 15 & R1, X1, X2 &  \\
\texttt{pytorch-model-cli} & 15 & R1, X1, O &  \\
\texttt{pytorch-model-recovery} & 15 & R1, X1 & \checkmark \\
\texttt{qemu-alpine-ssh} & 15 & R3, X2 &  \\
\texttt{qemu-startup} & 15 & R3, X2 &  \\
\texttt{query-optimize} & 15 & R1, X1, O & \checkmark \\
\texttt{raman-fitting} & 15 & R1 & \checkmark \\
\texttt{regex-chess} & 60 & R1, X1 &  \\
\texttt{regex-log} & 15 & R1, X1, O &  \\
\texttt{reshard-c4-data} & 60 & R1, R2, X1, O &  \\
\texttt{rstan-to-pystan} & 30 & R1 & \checkmark \\
\texttt{sam-cell-seg} & 120 & R1, X1, O &  \\
\texttt{sanitize-git-repo} & 15 & R1, R2 & \checkmark \\
\texttt{schemelike-metacircular-eval} & 40 & R1, X1 &  \\
\texttt{sparql-university} & 15 & R1, X1, O &  \\
\texttt{sqlite-db-truncate} & 15 & R1 & \checkmark \\
\texttt{sqlite-with-gcov} & 15 & R1, R2, X1 &  \\
\texttt{torch-pipeline-parallelism} & 15 & R1, X1 &  \\
\texttt{torch-tensor-parallelism} & 15 & R1, X1 &  \\
\texttt{train-fasttext} & 60 & R1, X1, O &  \\
\texttt{tune-mjcf} & 15 & R1, X1 & \checkmark \\
\texttt{video-processing} & 60 & R1, X1, O &  \\
\texttt{vulnerable-secret} & 15 & R1 & \checkmark \\
\texttt{winning-avg-corewars} & 60 & R1, X1 & \checkmark \\
\texttt{write-compressor} & 15 & R1, X1 & \checkmark \\
\bottomrule
\end{tabular}}

%% file: appendix/C_prices.tex
\section{Model Prices}
\label{app:prices}

Every dollar figure in the paper is model cost: the tokens a trial consumed, priced at the rates in Table~\ref{tab:prices}. Machine time is not counted. Prices are in dollars per million tokens as of September 2026.

Our own trials are priced at the AWS Bedrock rate of the model, read from its Bedrock model card (OpenAI models) or the public Bedrock price list (Claude, GLM). For GPT-5.6 Luna, Bedrock bills every uncached input token as a cache write, at 1.25 times the input rate; we do the same. The GPT-5.6 Luna + Codex CLI point of Figure~\ref{fig:intro} and Table~\ref{tab:acting-results} is priced at the AWS rate based on the token counts from Harbor Hub reports. We do not use the hub's own price for GPT-5.6 Luna, which is the rate before July 2026  (\$1.00 / \$0.10 / \$6.00) and five times higher than AWS.

Every other point of Figure~\ref{fig:intro} uses the cost Harbor Hub reports for that job. The hub's rates match the vendor's list price for GPT-5.5 + Codex, Opus 4.7 and Muse Spark 1.1; for Sonnet 5 + Claude Code the hub bills at Sonnet 4.6's rate, 1.5 times Sonnet 5's list price; for GPT-5.5 + Terminus-2 no fixed rate reproduces the hub's figure, which we use as reported. Claude Code trials that hit the time limit have no hub cost; we price their tokens at the rate the hub used for the rest of that arm.

\begin{table}[htb]
\caption{Prices in dollars per million tokens. \emph{Input}: uncached input; \emph{cached}: cache read; a dash means the price list has no cached rate and cached tokens are billed as input.}
\label{tab:prices}
\vspace{4pt}
\centering
\footnotesize
\setlength{\tabcolsep}{4pt}
\resizebox{\linewidth}{!}{\begin{tabular}{@{}llrrrl@{}}
\toprule
\textbf{Model} & \textbf{Used in} & \textbf{Input} & \textbf{Cached} & \textbf{Output} & \textbf{Source} \\
\midrule
GPT-5.6 Luna & all Luna arms; Luna + Codex & 0.20 (0.25 as a cache write) & 0.02 & 1.20 & Bedrock model card, global \\
Claude Sonnet 5 & critic, \Cref{app:crossmodel} & 2.00 & 0.20 & 10.00 & Anthropic list price \\
GLM-5 & critic, \Cref{app:crossmodel} & 1.00 & -- & 3.20 & Bedrock price list, us-east-1 \\
GPT-6 Astra & \Cref{app:stronger} & 13.75 (as cache write) & 1.10 & 55.00 & Bedrock model card, in-region \\
Claude Fable 5.1 & \Cref{app:stronger} & 13.75 (as cache write) & 0.275 & 55.00 & Bedrock price list, in-region \\
Claude Opus 5.5 & \Cref{app:stronger} & 5.50 (as cache write) & 0.22 & 22.00 & Bedrock price list, in-region \\
\addlinespace
GPT-5.5 + Codex & Figure~\ref{fig:intro} & 5.00 & 0.50 & 30.00 & Harbor Hub \\
Claude Opus 4.7 & Figure~\ref{fig:intro} & 6.25 (as cache write) & 0.50 & 25.00 & Harbor Hub \\
Muse Spark 1.1 & Figure~\ref{fig:intro} & 1.25 & 0.15 & 4.25 & Harbor Hub \\
Claude Sonnet 5 + Claude Code & Figure~\ref{fig:intro} & 3.75 & 0.30 & 15.00 & Harbor Hub (Sonnet 4.6 rate) \\
GPT-5.5 + Terminus-2 & Figure~\ref{fig:intro} & \multicolumn{3}{c}{hub total, no fixed rate} & Harbor Hub \\
\bottomrule
\end{tabular}}
\end{table}

%% file: appendix/D_statistical_methodology.tex
\section{Statistical Methodology}
\label{app:stats}

Every interval in the paper is a 95\% percentile bootstrap over tasks. Tasks are resampled with replacement 5000 times with a fixed seed; all trials, states and probes of a drawn task move together, and the statistic is recomputed on each resample. Differences between two arms are paired before resampling: by trial when both arms are read off the same trials (the baseline and the submitted state of Table~\ref{tab:acting-results}), and by state when two critics judge the same state (\S\ref{sec:q-act}, \S\ref{sec:q-info}). Sample sizes are given with each table and figure.

%% file: appendix/E_search_baselines.tex
\section{Search Baselines in Detail}
\label{app:search-baselines}

\input{figures/tab_seq_fork_numbers}
\input{figures/tab_mcts_numbers}
This appendix describes the two search baselines of \S\ref{sec:q-search}. Both use the actor of Table~\ref{tab:search-null}: GPT-5.6 Luna in the same harness with the same prompt, on all 89 tasks, five trials each, under the task's clock $T$. The baseline is that actor with no search: it runs until it first claims the task is done, and the verifier grades that state. Intervals are 95\% bootstrap intervals over tasks (\Cref{app:stats}).

\subsection{Branching Tools Without a Search Policy (seq\_fork)}
\label{app:search-fork}

This arm gives the actor the tool to branch. Its reply has two optional fields: \texttt{snapshot} saves the current environment under a label, and \texttt{restore} returns to a saved one. A restore replaces files, processes and shell state and keeps the conversation. The prompt says the tools may be used whenever the actor judges they help; it does not say when. There is no critic and no score. The trial ends when the actor claims completion or the clock runs out, and the verifier grades the live environment.

Pass@1 is \SeqForkPass{} against \SeqPass{} for the baseline ($\Delta =$~\SeqForkDeltaSeq\,\SeqForkDeltaSeqCI; Table~\ref{tab:seq_vs_seq_fork}). The actor touched the tools in only \SeqForkToolTrials{} of \SeqForkNTrials{} trials: it saved \SeqForkAgentSnapshots{} snapshot and restored \SeqForkRestores{} times. Two restores, in \textsc{db-wal-recovery}, went back to $\Sinit$ to recover a write-ahead log the actor had destroyed by opening the database. The third, in \textsc{path-tracing}, returned to a saved renderer after a change made it worse. Given the tools without a policy, the actor does not explore alternatives or protect its progress before a risky step; it restores only to undo damage already done. (\SeqForkNLost{} trials in which the actor's commands destroyed the sandbox before grading are dropped.)

\begin{table}[!ht]
\caption{The actor with and without snapshot and restore tools, 89 tasks of Terminal-Bench~2.1, five trials per task.}
\label{tab:seq_vs_seq_fork}
\vspace{4pt}

\centering
\small
\begin{tabular}{lcc}
\toprule
 & Baseline & + snapshot and restore tools \\
\midrule
Pass@1 (passed / graded trials) & \SeqPass{} (\SeqNPassed/\SeqNGraded) & \SeqForkPass{} (\SeqForkNPassed/\SeqForkNGraded) \\
Trials using a tool & -- & \SeqForkToolTrials \\
Snapshots saved by the actor & -- & \SeqForkAgentSnapshots \\
Restores (to $\Sinit$) & -- & \SeqForkRestores{} (\SeqForkRestoresSzero) \\
Model cost / sweep & \SeqCost & \SeqForkCost \\
\bottomrule
\end{tabular}
\end{table}

\subsection{Tree Search With a Trajectory-Only Critic (MCTS)}
\label{app:search-mcts}

This arm adds a search policy and a critic to the actor. It builds one search tree per trial, and each node of the tree is a snapshot of the environment.

To expand a node, the search restores the node's snapshot, lets the actor run five turns, and saves the result as a new child node. A node has at most two children.

The critic is the trajectory-only critic $C^{\mathrm{traj}}$ (\Cref{app:prompts}). It sees the task and the actor's turns on the path to a node. It scores each new node once, from 0 to 100. A node's value is the mean score of all nodes in its subtree.

To choose the next node to expand, the search starts at the root. If the current node has fewer than two children, the search expands it. Otherwise it moves to the child with the highest UCT score~\citep{kocsis2006uct}, with $c_p = 0.12$, and repeats. A node at which the actor declared the task finished is a candidate for submission, and the search never expands it.

The search stops when no node can be expanded or when the clock $T$ runs out. Snapshots, restores and critic calls all count against $T$. The search then submits the candidate with the highest value, or the node with the highest value if there is no candidate. The verifier grades the submitted state.

Pass@1 is \MctsPass{} for MCTS and \SeqPass{} for the baseline ($\Delta =$~\MctsDeltaSeq\,\MctsDeltaSeqCI; Table~\ref{tab:mcts-null}). The interval contains zero. MCTS costs \MctsCostRatio$\times$ as much as the baseline, and the critic accounts for \MctsCriticCostShare{} of that cost.

\begin{table}[!ht]
\caption{Pass@1 of the baseline and MCTS, five trials per task. MCTS has no graded trial on \textsc{kv-store-grpc}, so its pass@1 is over \MctsNGradedTasks{} tasks and the baseline's over 89. $\Delta$ is MCTS minus the baseline; its interval is a bootstrap over tasks that draws the same tasks for both arms (\Cref{app:stats}).}
\label{tab:mcts-null}
\vspace{4pt}
\centering
\small
\begin{tabular}{lcc}
\toprule
 & Pass@1 & Model cost / sweep \\
\midrule
Baseline   & \SeqPass & \SeqCost \\
MCTS       & \MctsPass & \MctsCost \\
\midrule
$\Delta$ [95\% CI] & \MctsDeltaSeq\,\MctsDeltaSeqCI & \MctsCostRatio$\times$ \\
\bottomrule
\end{tabular}
\end{table}

The search did choose between paths. It opened a second branch in \MctsBranched{} of trials. In \MctsLaterBranch{} of trials, the submitted state came from a branch opened after the first path. In \MctsSaturated{} of trials, the search ended before the clock ran out, because every leaf was a state the actor had declared finished. 

We drop \MctsNLostCkpt{} trials lost to a failed snapshot, and the \MctsNNoGrade{} trials of \textsc{kv-store-grpc}, whose verifier could not run.

%% file: appendix/F_critic_by_position.tex
\section{Critic Metrics by Position}
\label{app:critic-by-position}

Table~\ref{tab:critic-by-position} gives the numbers behind Figure~\ref{fig:critic-by-position}. It also gives the false-accept and false-reject rates at the intermediate depths, two checks of the ranking, and the mean absolute error (MAE).

All critics see the same task, the same trajectory and the same remaining budget. \textsc{read} and \textsc{act} can also run commands in a fork of the state: \textsc{read} only read-only commands, \textsc{act} any command (\S\ref{sec:q-act}).

At an intermediate state, the outcome is the pass rate of the \CbpSamples{} continuations from that state. A continuation that the verifier could not grade is left out of this rate. We exclude \CbpNMidNoLabel{} intermediate and \CbpNFinalNoLabel{} final states that have no graded outcome. As in Table~\ref{tab:access-class}, we also exclude the \CbpNFinalParseFail{} final state on which the verdict of $C^{\mathrm{read}}$ could not be parsed.

\paragraph{Label noise.} A pass rate from \CbpSamples{} continuations is a noisy label. Its standard error is at most \CbpNoiseSEMax{} points, and \CbpNoiseSE{} points in root mean square over the intermediate states. The critics never see the continuations, so this noise can flip the order of a pair but favors no critic: it pulls pairwise accuracy towards 50\%. For \CbpNoiseDecidedShare{} of the intermediate states, the continuations all pass or all fail. On the \CbpNDecPairsPooled{} decided pairs, where one state's continuations all pass and the other's all fail, the critics order \CbpPairDecPooledLo--\CbpPairDecPooledHi{} of the pairs correctly. The gain of $C^{\mathrm{act}}$ over $C^{\mathrm{traj}}$ on these pairs is \CbpPairDecGapActTrajPooled.

\paragraph{Choosing a state.} Search uses the critic to choose among states. Among the states of a task at one depth, the state the critic rates highest has a continuation pass rate \CbpSelPooledLo--\CbpSelPooledHi{} points above the mean of those states, which is what a random choice gets. This gain appears only late in the trajectory: at 25\% depth it is $-$1.6 points\,[$-$6.7,\,$+$3.7] for \textsc{traj}, $+$2.8\,[$-$1.4,\,$+$6.9] for \textsc{read} and $+$0.2\,[$-$5.3,\,$+$5.4] for \textsc{act}, every interval including zero; at 75\% depth it is $+$11.0 to $+$13.6 points, every interval above zero (Table~\ref{tab:critic-by-position}). The critics therefore carry some signal at intermediate states, but access adds little to it: the gain of $C^{\mathrm{act}}$ over $C^{\mathrm{traj}}$ is \CbpSelGapActTrajPooled.

\paragraph{Access at the two positions.} Each gain from access compares two critics on the same states and outcomes. In pairwise accuracy, the gain of $C^{\mathrm{act}}$ over $C^{\mathrm{traj}}$ is \CbpPairGapActTrajFinal{} at the final state and \CbpPairGapActTrajPooled{} at the intermediate depths pooled. The difference between the two gains is not significant (\CbpPairAccessDiffActTraj).

The intervals in Figure~\ref{fig:critic-by-position} are 95\% bootstrap intervals over tasks (\Cref{app:stats}).

\input{figures/tab_critic_by_position}

%% file: figures/tab_critic_by_position.tex
\input{figures/fig_critic_by_position_numbers}
\begin{table}[tp]
\caption{The numbers behind Figure~\ref{fig:critic-by-position}. At the intermediate depths, the critic's output is its estimate $\hat q$ and the outcome is the pass rate of \CbpSamples{} continuations. At the final state, the output is the verdict $\hat v \in \{0, 1\}$ and the outcome is the verifier's judgment. A pair is two states of the same task at the same depth (at the final state: two final states of the same task) whose outcomes differ; pairwise accuracy is the share of pairs that the output orders like the outcome, with tied outputs counting one half. A decided pair has outcomes 0 and 1. The selection gain is the outcome of the state with the highest output among a task's states at one depth, minus the mean outcome of those states. A false accept is a failing state with output $\ge$\,\CbpAcceptAt; a false reject is a passing state with output $<$\,\CbpAcceptAt. At intermediate depths, these two rates use only the states from which all continuations pass or all fail. MAE is the mean of $|\text{output} - \text{outcome}|$ (at the final state, the misclassification rate); \emph{best constant} is the MAE of the best single answer. Rows $X-Y$ are differences between two critics. The column \emph{all} pools the three depths.}
\label{tab:critic-by-position}
\vspace{4pt}
\centering
\small
\input{figures/tab_critic_by_position_body}
\end{table}

%% file: figures/tab_critic_by_position_body.tex
\begin{tabular}{@{}llrrr@{\hskip 1em}r@{\hskip 1.4em}r@{}}
\toprule
 & & \multicolumn{4}{c}{intermediate state, depth} & \\
\cmidrule(lr){3-6}
 & & 25\% & 50\% & 75\% & all & final state \\
\midrule
Pairwise accuracy (\%) $\uparrow$ & \textsc{traj} & 49 & 56 & 64 & 54 & 65 \\
 & \textsc{read} & 56 & 55 & 69 & 59 & 61 \\
 & \textsc{act} & 52 & 62 & 68 & 59 & 76 \\
\addlinespace
Pairwise, decided pairs (\%) $\uparrow$ & \textsc{traj} & 56 & 73 & 68 & 68 & 65 \\
 & \textsc{read} & 67 & 70 & 78 & 74 & 61 \\
 & \textsc{act} & 53 & 62 & 80 & 71 & 76 \\
\addlinespace
Selection gain (points) $\uparrow$ & \textsc{traj} & $-$2 & $+$7 & $+$11 & $+$4 & $+$14 \\
 & \textsc{read} & $+$3 & $+$3 & $+$14 & $+$6 & $+$8 \\
 & \textsc{act} & 0 & $+$8 & $+$13 & $+$6 & $+$22 \\
\addlinespace
False reject (\%) $\downarrow$ & \textsc{traj} & 2 & 1 & 1 & 1 & 60 \\
 & \textsc{read} & 28 & 11 & 4 & 14 & 52 \\
 & \textsc{act} & 8 & 6 & 2 & 5 & 22 \\
\addlinespace
False accept (\%) $\downarrow$ & \textsc{traj} & 91 & 81 & 80 & 84 & 5 \\
 & \textsc{read} & 54 & 54 & 61 & 57 & 10 \\
 & \textsc{act} & 75 & 66 & 59 & 66 & 22 \\
\addlinespace
MAE (\%) $\downarrow$ & \textsc{traj} & 34 & 32 & 30 & 32 & 39 \\
 & \textsc{read} & 34 & 32 & 28 & 32 & 36 \\
 & \textsc{act} & 32 & 31 & 26 & 30 & 22 \\
 & \textsc{read}$-$\textsc{traj} & 0 & 0 & $-$1 & 0 & $-$3 \\
 & \textsc{act}$-$\textsc{traj} & $-$2 & $-$1 & $-$4 & $-$2 & $-$17 \\
 & \textsc{act}$-$\textsc{read} & $-$2 & $-$1 & $-$2 & $-$2 & $-$14 \\
 & best constant & 36 & 38 & 37 & 38 & 38 \\
 & \textsc{traj}, summary + ledger & 34 & 32 & 30 & 32 & 37 \\
 & \textsc{traj}, checklist & 33 & 33 & 30 & 32 & 41 \\
\midrule
States & & 404 & 333 & 374 & 1111 & 406 \\
Tasks & & 88 & 83 & 85 & 88 & 89 \\
Pairs & & 317 & 184 & 171 & 672 & 144 \\
Decided pairs & & 18 & 30 & 69 & 117 & 144 \\
\bottomrule
\end{tabular}

%% file: appendix/G_critic_cost.tex
\section{Critic Cost}
\label{app:critic-cost}
\input{figures/fig_critic_cost_numbers}

Figure~\ref{fig:critic-cost} gives the model cost of each critic per state, at the prices of \Cref{app:prices}.

At intermediate states, $C^{\mathrm{traj}}$ costs \CcostTrajMid{} cents per state, $C^{\mathrm{read}}$ \CcostReadMid{} cents and $C^{\mathrm{act}}$ \CcostActMid{} cents.

At candidate final states, $C^{\mathrm{act}}$ costs \CcostActFinal{} cents per state. This is \CcostActOverTrajFinal$\times$ the cost of $C^{\mathrm{traj}}$ and \CcostActOverReadFinal$\times$ the cost of $C^{\mathrm{read}}$.

$C^{\mathrm{act}}$ costs \CcostActFinalOverMid$\times$ as much at a final state as at an intermediate state, because it takes more turns there: \CcostActTurnsFinal{} on average, against \CcostActTurnsMid.

\input{figures/fig_critic_cost}

%% file: figures/fig_critic_cost_numbers.tex
\def\CcostNMid{1111}
\def\CcostNFinal{406}
\def\CcostTrajMid{0.32}
\def\CcostTrajFinal{0.36}
\def\CcostReadMid{0.51}
\def\CcostReadFinal{0.85}
\def\CcostActMid{0.59}
\def\CcostActFinal{1.11}
\def\CcostReadTurnsMid{3.1}
\def\CcostReadTurnsFinal{4.1}
\def\CcostActTurnsMid{3.5}
\def\CcostActTurnsFinal{4.6}
\def\CcostActOverTrajMid{1.9}
\def\CcostActOverTrajFinal{3.1}
\def\CcostActOverReadMid{1.2}
\def\CcostActOverReadFinal{1.3}
\def\CcostActFinalOverMid{1.9}

%% file: figures/fig_critic_cost.tex
\begin{figure}[!htb]
\centering
\includegraphics[width=\linewidth]{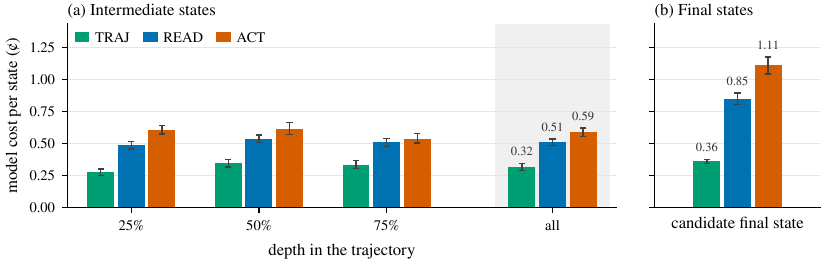}
\caption{Model cost per state of the critics of Figure~\ref{fig:critic-by-position}, on the same states. \textbf{(a)}~Intermediate states at 25/50/75\% depth; \emph{all} pools the three depths. \textbf{(b)}~Candidate final states. Intervals are 95\% bootstrap intervals over tasks.}
\label{fig:critic-cost}
\end{figure}

%% file: appendix/H_traj_prompts.tex
\section{Trajectory-Only Critic Prompts}
\label{app:traj-variants}

Figure~\ref{fig:traj-variants} compares three prompts for the trajectory-only critic (\Cref{app:prompts}) on the states of Figure~\ref{fig:critic-by-position}. None of these critics runs commands.

The \emph{plain} prompt is the one that \textsc{traj} uses in the main text. \emph{Summary + ledger} adds a summary of the trajectory and a per-step ledger of facts. \emph{Checklist} adds a list of failure modes to look for.

No variant differs significantly from \textsc{traj} at any position. At every intermediate depth, the three are within \CbpTrajVarSpreadMid{} of one another in MAE.

\input{figures/fig_traj_variants}

%% file: figures/fig_traj_variants.tex
\begin{figure}[!htb]
\centering
\includegraphics[width=\linewidth]{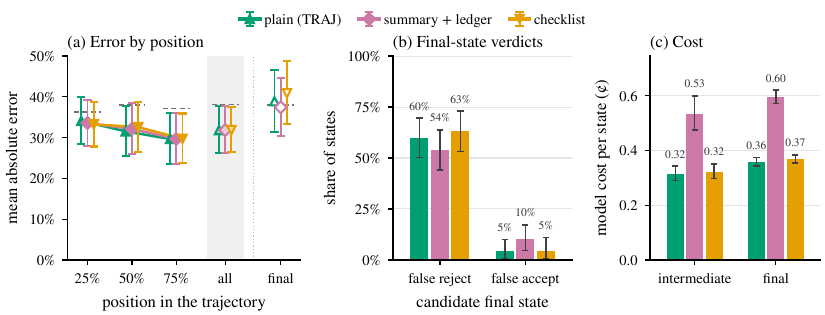}
\caption{The three trajectory-only prompts on the states of Figure~\ref{fig:critic-by-position}; \emph{plain} is \textsc{traj} in Figure~\ref{fig:critic-by-position}. \textbf{(a)}~Mean absolute error by position (Table~\ref{tab:critic-by-position}); dashes mark the best constant answer. \textbf{(b)}~At the candidate final state: the share of passing states rejected and the share of failing states accepted. \textbf{(c)}~Model cost per state (\Cref{app:prices}); for \emph{summary + ledger} it includes the calls that write the summary and the ledger. Intervals are 95\% bootstrap intervals over tasks.}
\label{fig:traj-variants}
\end{figure}

%% file: appendix/I_s0_reference.tex
\section{The Reference-State Critic: Protocol}
\label{app:s0-reference}

\input{figures/fig_s0_reference_numbers}
This appendix describes how the comparison of \S\ref{sec:q-info} was run. The snapshots of the final states of \S\ref{sec:q-act} were not kept. We rebuilt each state in a fresh environment by replaying, in order, every command the actor had issued. Of the \SzrNRecorded{} final states, we use \SzrNKept. We drop the other \SzrNDropped{} because the replay, the verifier or the fork of $\Sinit$ failed on them. To check the replay, we ran the verifier on each rebuilt state and compared its result with the one recorded in the original trial. The two results are the same on \SzrReplayMatch{} states and differ on \SzrReplayMismatch; the other \SzrNNoRecorded{} states have no recorded result to compare with. In the comparison, whether a state passes is decided by the verifier's result on the rebuilt state. Each critic judged every state twice. $C^{\mathrm{act}}$ uses the same prompt as the critic of \S\ref{sec:q-act}. The prompt of $C^{\mathrm{act}}{+}\Sinit$ adds one paragraph. It tells the critic that there is a second environment, how to run a command in it, and how to copy a file from it.

%% file: appendix/J_other_models.tex
\section{Different Critic Models}
\label{app:crossmodel}

The critic does not have to use the actor's model. We test two other models as the critic. The actor stays GPT-5.6 Luna, and nothing else changes: the same prompts, the same clock $T$ and the same revision budget, on the same \LoopXSonnetNTasks{} tasks.

\paragraph{Claude Sonnet 5.}
Sonnet~5 is about as strong as Luna. As an agent in Claude Code, it passes \LoopSonnetAlonePassCI{} of the trials on these tasks, against \LoopCodexPassCI{} for Luna in the Codex CLI.

The critic acts only after the first claim, so the first claims pass at about the same rate as with the Luna critic: \LoopXSonnetRoundZero{} against \LoopRoundZero. The loop then submits a passing state in \LoopXSonnetHandIn{} of the trials. The gain is \LoopXSonnetGainHandIn, as large as with the Luna critic (\LoopGainHandIn; Figure~\ref{fig:acting-crossmodel}a). The per-task pass rate rises on \LoopXSonnetTasksUp{} tasks and falls on \LoopXSonnetTasksDown.

\input{figures/fig_acting_crossmodel}

The gain comes about in the same way as with Luna (\S\ref{sec:acting-mechanism}). The Sonnet critic rejects \LoopXSonnetCatchRate{} of the failing first claims (Luna: \LoopCatchRate), and the actor turns \LoopXSonnetRescued{} of them into passing states. False rejections cost \LoopXSonnetFRLost{} passing first claims.

Sonnet's probes take longer: a median of \LoopXSonnetProbeSec\,s and \LoopXSonnetProbeCmds{} commands, against \LoopProbeSec\,s and \LoopProbeCmds{} for Luna (Figure~\ref{fig:acting-crossmodel}b). Because the clock is shared, \LoopXSonnetClockCritic{} loops run out of time during a probe, against \LoopClockCritic{} with Luna. Each of them submits its last proposed state.

\paragraph{GLM-5.}
We also use Z.AI's GLM-5 as the critic, with the same setup. The first claims pass at about the same rate, \LoopXGlmRoundZero{} against \LoopRoundZero. The loop submits a passing state in \LoopXGlmHandIn{} of the trials, a gain of \LoopXGlmGainHandIn{} (Figure~\ref{fig:acting-crossmodel}a). The per-task pass rate rises on \LoopXGlmTasksUp{} tasks and falls on \LoopXGlmTasksDown. The gain is smaller than with the Luna critic, but the difference, \LoopXGlmGainVsMain, is not significant.

GLM-5 is a more lenient critic than Luna. It rejects only \LoopXGlmCatchRate{} of the failing first claims (Luna: \LoopCatchRate), so fewer of them go back to the actor, which turns \LoopXGlmRescued{} into passing states. It also rejects fewer passing first claims, \LoopXGlmFirstFR{} against \LoopFirstFR, so false rejections cost only \LoopXGlmFRLost{} passing first claim. Its probes take a median of \LoopXGlmProbeSec\,s, between Luna's and Sonnet's (Figure~\ref{fig:acting-crossmodel}b), and \LoopXGlmProbeCmds{} commands.

%% file: figures/fig_acting_crossmodel.tex
\begin{figure}[!ht]  %
\centering
\includegraphics[width=\linewidth]{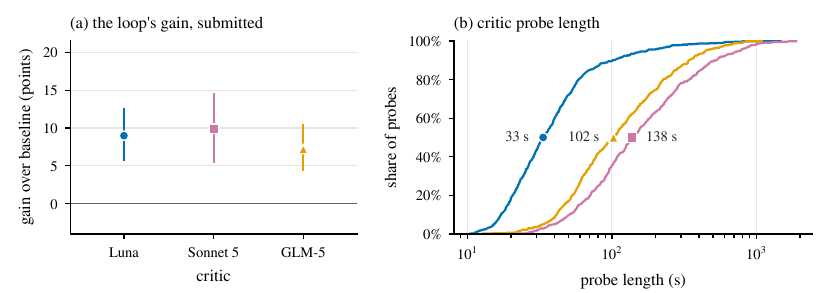}
\caption{GPT-5.6 Luna as the actor, with each model as the critic. (a)~The loop's gain over the baseline in the submitted pass rate (Table~\ref{tab:acting-results}); bars are 95\% bootstrap intervals over tasks. (b)~How long the critic's probes take, over the probes that return a verdict; the markers are the medians.}
\label{fig:acting-crossmodel}
\end{figure}

%% file: appendix/K_restart.tex
\section{Resuming or Restarting}
\label{app:restart}

\input{figures/tab_restart_numbers}
After a rejection, \sys{} resumes the actor at the state it claimed (\S\ref{sec:acting-design}). The alternative is to restart. A restart resets the environment and the actor's conversation to $\Sinit$, and the actor starts the task again, with the critic's findings added after the task instruction.

We run this restart variant on all \RstNTasks{} tasks of Terminal-Bench~2.1, five trials each. Nothing else changes: the same model is the actor and the critic, the critic and its feedback are the same, a trial has at most six rounds, and the actor and the critic share the time limit $T$. Environment failures lost \RstNLost{} trials, which leaves \RstNRuns.

Table~\ref{tab:restart} compares the two variants; the \sys{} row repeats Table~\ref{tab:acting-results}. Restarting submits a passing state in \RstHandInPct{} of trials, against \LoopHandInPct{} for resuming. The difference is \RstDiffHandInPts{} points \RstDiffHandInCIPts.

\input{figures/tab_restart}

Restarting also costs \RstCostRatio{} times as much in model calls, and a trial takes \RstTimeMin{} minutes on average instead of \RstResumeTimeMin. The critic's time is the same in both variants, so the extra time is the actor's: after a restart, the actor has to redo the work of the rejected attempt before it can act on the critic's findings. Of the \RstRunsRestarted{} trials that restarted at least once, \RstRestartedOutOfTime{} ran out of time before the loop ended.

%% file: figures/tab_restart.tex
\begin{table}[!ht]
\caption{Resuming vs.\ restarting after a rejection, on the \RstNTasks{} tasks of Terminal-Bench~2.1, five trials per task. \emph{Pass rate} and cost as in Table~\ref{tab:acting-results}; time is the actor's and the critic's per trial, averaged.}
\label{tab:restart}
\vspace{4pt}
\centering
\small
\setlength{\tabcolsep}{5pt}
\input{figures/tab_restart_body}
\end{table}

%% file: figures/tab_restart_body.tex
\begin{tabular}{@{}lccc@{}}
 & \textbf{Pass rate} & \textbf{Total model cost} & \textbf{Time / trial} \\
\sys{} & \textbf{72.4\%} & \$23.14 & 7.8 min \\
Restart from $\Sinit$ & 66.3\% & \$38.12 & 9.3 min \\
\end{tabular}

%% file: appendix/L_stronger_models.tex
\section{Stronger Models as Actor and Critic}
\label{app:stronger}

Table~\ref{tab:acting-results} uses GPT-5.6 Luna, which passes \LoopRoundZeroPct{} of its first claims on Terminal-Bench~2.1. Here we repeat the loop with stronger models: GPT-6 Astra and Claude Opus~5.5 as the actor, and Claude Fable~5.1 or Opus~5.5 as the critic. On Harbor Hub, Astra in the Codex CLI passes \StrongAstraHubPct{} of 25 trials per task. We use three combinations of actor and critic (Table~\ref{tab:stronger-models}).

We run them on \StrongNTasks{} tasks, chosen from the 31 tasks that Astra fails at least once in its 25 hub trials:
\begin{itemize}
\item ten on which its first claim passes in some of our trials and fails in others, so that the loop has something to recover;
\item four that it always passes, as controls for false rejections;
\item one that it never passes, as a control for false accepts.
\end{itemize}
The other settings are the same: the actor, the critic and the snapshots share the clock $T$, and a trial has at most six claims and 50 actor turns after the first rejection.

\input{figures/tab_stronger_models}

None of the three combinations gains significantly:
\begin{itemize}
\item \emph{Astra actor, Fable~5.1 critic.} The submitted pass rate stays at the baseline: \StrongAstraFableHandInPct{} against \StrongAstraFableRoundZeroPct, $\Delta = $ \StrongAstraFableGainPts{} points \StrongAstraFableGainCIPts.
\item \emph{Astra actor, Opus~5.5 critic.} The pass rate rises by \StrongAstraOpusGainPts{} points \StrongAstraOpusGainCIPts, an interval that includes zero. \StrongAstraOpusRunsUp{} trials gain a passing state and \StrongAstraOpusRunsDown{} loses one.
\item \emph{Opus~5.5 on both sides.} The loop changes nothing. The critic accepts almost every first claim, no trial is rescued or lost, and the submitted rate equals the baseline (\StrongOpusOpusHandInPct).
\end{itemize}
The mechanism of \S\ref{sec:acting-mechanism} explains this. With Luna, only \StrongLunaLunaRoundZeroPct{} of first claims on these tasks pass, and the loop works on the rest. With these actors, most first claims already pass, so there is little to recover. The remaining failures are ones that a critic of the same strength does not catch either.

One issue comes from the model, not from the loop. Claude's content classifier answers two life-science tasks with empty replies, so \StrongOpusOpusRefused{} of the Opus~5.5 actor trials are refused; they count as 0 (column \emph{Refused}). These trials ran to the clock on empty replies: the harness now stops an actor after six empty replies, but these trials ran before that change. Astra's Bedrock endpoint rejects some security tasks with a policy error, but no Astra trial on these tasks was refused.

%% file: figures/tab_stronger_models.tex
\begin{table}[!htb]
\caption{\sys{} with stronger models as the actor and the critic, on the same \StrongNTasks{} tasks of Terminal-Bench~2.1, five trials per task. Luna: GPT-5.6 Luna (the trials of Table~\ref{tab:acting-results} on these tasks); Astra: GPT-6 Astra; Fable, Opus: Claude Fable~5.1, Claude Opus~5.5. \emph{Baseline} and \sys{} as in Table~\ref{tab:acting-results}. Intervals are 95\% bootstrap intervals over tasks. \emph{Refused}: trials the actor's model refused; they count 0 in both columns. \emph{Cost}: the total over the trials that were not refused (\Cref{app:prices}).}
\label{tab:stronger-models}
\vspace{4pt}
\centering
\small
\setlength{\tabcolsep}{4pt}
\resizebox{\linewidth}{!}{\input{figures/tab_stronger_models_body}}
\end{table}

%% file: figures/tab_stronger_models_body.tex
\begin{tabular}{@{}lrrcccrr@{}}
\textbf{Actor / Critic} & \textbf{Tasks} & \textbf{Trials} & \textbf{Baseline} & \textbf{\sys{}} & \textbf{$\Delta$ [95\% CI]} & \textbf{Refused} & \textbf{Cost} \\
Luna / Luna & 15 & 75 & 33.3\% & \textbf{40.0\%} & $+$6.7\,[$-$2.7,\,$+$16.0] & 0 & \$4.04 \\
Astra / Fable~5.1 & 15 & 75 & 61.3\% & \textbf{62.7\%} & $+$1.3\,[$-$6.7,\,$+$9.3] & 0 & \$126.94 \\
Astra / Opus~5.5 & 15 & 75 & 69.3\% & \textbf{73.3\%} & $+$4.0\,[$-$2.7,\,$+$10.7] & 0 & \$65.83 \\
Opus~5.5 / Opus~5.5 & 15 & 75 & 69.3\% & 69.3\% & 0.0\,[0.0,\,0.0] & 8 & \$52.80 \\
\end{tabular}

%% file: appendix/M_swebench_pro.tex
\section{SWE-bench Pro: Tasks and Grading}
\label{app:swepro-setup}

This appendix gives the setup of \S\ref{app:swepro}. We use the public split of SWE-bench Pro V2.0.0, released on 22 September 2026, with the benchmark's own task images. The loop is the one of \S\ref{sec:acting-design} with the same prompts and GPT-5.6 Luna as actor and critic. The actor, the critic and every snapshot, restore and fork share the task's 50-minute limit. A trial has at most six claims and 50 actor turns after the first rejection.

\paragraph{The two task sets.}
Both sets are fixed by files released with the benchmark. Neither depends on how GPT-5.6 Luna or \sys does on any task.
\begin{itemize}
\item \emph{Hard} is the benchmark's HARD-51 list: the tasks that at least two of five frontier model families fail under V2's protocol. We leave out its one NodeBB task, which gives \SweProNHard{} tasks.
\item \emph{Medium} comes from the per-task results of \SweProNReleasedRuns{} runs that the benchmark's repository released: 2025 models (Claude Sonnet~4.5, Claude Sonnet~4 twice, Claude Opus~4.1, Gemini~2.5 Pro Preview, GPT-4o, GPT-5, gpt-oss and Kimi~K2) on the first release. Of the \SweProNVTwo{} V2 tasks, \SweProNSolvedByReleased{} are solved by at least one of these runs. Medium is every task that none of them solves, outside HARD-51 and NodeBB: \SweProNMedium{} tasks. The benchmark itself has no medium tier; the name is ours. The 2025 models failed these tasks, while at most one of the five model families behind HARD-51 fails each of them.
\end{itemize}
The task lists and the released result files come with our code, and our script rebuilds both sets from the released files.

We leave out all \SweProNNodeBB{} NodeBB tasks because their reference patches fail in our environment even with network access.

\paragraph{Tasks we do not score.}
A task is scored only if its reference patch passes in the same environment as our trials. This rule does not look at our trials. \SweProNExcluded{} tasks fail it (\SweProNExcludedHard{} hard, \SweProNExcludedMedium{} medium). In all of them the tests need the network while grading: \SweProNExclGo{} Go tasks download modules, \SweProNExclNode{} task downloads Node.js and \SweProNExclNet{} Python test opens a connection when it is loaded. V2's protocol gives the verifier network access. Our sandbox allows no traffic except to the model endpoint, for grading as well as for the agent, so these tasks cannot be graded here.

\paragraph{Trials.}
We report the first trial of each of the \SweProNTasks{} scored tasks, so each task counts once in both sets. The hard tasks were run four more times; Table~\ref{tab:swepro-tiers} reports all five. No trial, in either set, ended without a claim.

\begin{table}[!htb]
\caption{SWE-bench Pro by task set. Baseline: the first proposed state; \sys: the submitted state. Gains are in points, with 95\% intervals over tasks (\Cref{app:stats}).}
\label{tab:swepro-tiers}
\vspace{4pt}
\centering
\small
\begin{tabular}{@{}lrrccc@{}}
\toprule
 & Tasks & Trials & Baseline & \sys & Gain [95\% CI] \\
\midrule
Hard, first trial & \SweProNHardScored & \SweProHardRuns & \SweProHardFirstPct & \SweProHardSubmittedPct & \SweProHardGain \\
Hard, all five trials & \SweProNHardScored & \SweProHardAllRuns & \SweProHardAllFirstPct & \SweProHardAllSubmittedPct & \SweProHardAllGain \\
Medium & \SweProNMediumScored & \SweProMediumRuns & \SweProMediumFirstPct & \SweProMediumSubmittedPct & \SweProMediumGain \\
\midrule
Both, first trial (Figure~\ref{fig:swepro}) & \SweProNTasks & \SweProNTasks & \SweProFirstPctOne & \SweProSubmittedPctOne & \SweProGainTable \\
\bottomrule
\end{tabular}
\end{table}

\paragraph{Grading.}
V2's locked protocol keeps the agent offline except for the model endpoint. It then captures the agent's \texttt{git diff}, including new files, applies it to a fresh copy of the image and runs the verifier there. Our agent phase is offline in the same way. We change one line of each task file, from no network to a public network, because Waypoint cannot create a sandbox without a network; a network namespace around the whole trial then lets traffic reach only the model endpoint. At the start of every trial, a check from inside the namespace confirms that the endpoint connects and that GitHub, PyPI, the Go module proxy, npm and direct connections by address are refused. Each trial also saved its DNS and firewall log; in all \SweProNEgressRuns{} trials we report, the firewall admitted only addresses of the model endpoint.

We grade differently. Each proposed state is graded by the task's unchanged V2 verifier on a fork of that state. The verifier first resets the repository's test code: it restores tracked test files and fixtures, deletes test files the agent added and removes cached bytecode. It then applies the hidden tests and runs them. The fork is discarded after grading, so the hidden tests never reach the actor's or the critic's sandbox. The loop records the grade but never reads it; only the critic's verdict decides whether the actor continues.

The difference from the protocol is what the graded state contains. A patch carries only the repository's tracked and new files. It drops files outside the repository or ignored by git, installed packages and running processes. Our fork keeps them. With no network, no trial could download a package, but a local change outside the repository would count in our grading and not in the protocol's.

The gain comes from the medium tasks. On the hard tasks it cannot be told apart from zero, with the first trial or with all five (Table~\ref{tab:swepro-tiers}).

%% file: main.bbl
\begin{thebibliography}{43}
\providecommand{\natexlab}[1]{#1}
\providecommand{\url}[1]{\texttt{#1}}
\expandafter\ifx\csname urlstyle\endcsname\relax
  \providecommand{\doi}[1]{doi: #1}\else
  \providecommand{\doi}{doi: \begingroup \urlstyle{rm}\Url}\fi

\bibitem[Antoniades et~al.(2024)Antoniades, {\"O}rwall, Zhang, Xie, Goyal, and
  Wang]{antoniades2024swesearch}
Antonis Antoniades, Albert {\"O}rwall, Kexun Zhang, Yuxi Xie, Anirudh Goyal,
  and William Wang.
\newblock {SWE-Search}: Enhancing software agents with {Monte Carlo} tree
  search and iterative refinement.
\newblock \emph{arXiv preprint arXiv:2410.20285}, 2024.
\newblock URL \url{https://arxiv.org/abs/2410.20285}.

\bibitem[Brown et~al.(2024)Brown, Juravsky, Ehrlich, Clark, Le, Ré, and
  Mirhoseini]{brown2024large}
Bradley Brown, Jordan Juravsky, Ryan Ehrlich, Ronald Clark, Quoc~V. Le,
  Christopher Ré, and Azalia Mirhoseini.
\newblock Large language monkeys: Scaling inference compute with repeated
  sampling, 2024.
\newblock URL \url{https://arxiv.org/abs/2407.21787}.

\bibitem[Browne et~al.(2012)Browne, Powley, Whitehouse, Lucas, Cowling,
  Rohlfshagen, Tavener, Perez, Samothrakis, and Colton]{browne2012survey}
Cameron~B. Browne, Edward Powley, Daniel Whitehouse, Simon~M. Lucas, Peter~I.
  Cowling, Philipp Rohlfshagen, Stephen Tavener, Diego Perez, Spyridon
  Samothrakis, and Simon Colton.
\newblock A survey of {Monte Carlo} tree search methods.
\newblock \emph{IEEE Transactions on Computational Intelligence and AI in
  Games}, 4\penalty0 (1):\penalty0 1--43, 2012.
\newblock \doi{10.1109/TCIAIG.2012.2186810}.

\bibitem[Chen et~al.(2025)Chen, Pan, Clark, Su, Zheutlin, Bhavya, Arora, Deng,
  Jha, and Xu]{chen2025stratus}
Yinfang Chen, Jiaqi Pan, Jackson Clark, Yiming Su, Noah Zheutlin, Bhavya
  Bhavya, Rohan~R. Arora, Yu~Deng, Saurabh Jha, and Tianyin Xu.
\newblock {STRATUS}: A multi-agent system for autonomous reliability
  engineering of modern clouds.
\newblock In \emph{Advances in Neural Information Processing Systems},
  volume~38, 2025.
\newblock \doi{10.52202/085713-1671}.
\newblock URL
  \url{https://papers.nips.cc/paper_files/paper/2025/hash/47a6e9e2c3019f13ad94a0f259fe4970-Abstract-Conference.html}.

\bibitem[Cobbe et~al.(2021)Cobbe, Kosaraju, Bavarian, Chen, Jun, Kaiser,
  Plappert, Tworek, Hilton, Nakano, Hesse, and Schulman]{cobbe2021training}
Karl Cobbe, Vineet Kosaraju, Mohammad Bavarian, Mark Chen, Heewoo Jun, Lukasz
  Kaiser, Matthias Plappert, Jerry Tworek, Jacob Hilton, Reiichiro Nakano,
  Christopher Hesse, and John Schulman.
\newblock Training verifiers to solve math word problems, 2021.
\newblock URL \url{https://arxiv.org/abs/2110.14168}.

\bibitem[Coulom(2007)]{coulom2006efficient}
R{\'e}mi Coulom.
\newblock Efficient selectivity and backup operators in {Monte-Carlo} tree
  search.
\newblock In H.~Jaap van~den Herik, Paolo Ciancarini, and H.~H. L. M.~(Jeroen)
  Donkers (eds.), \emph{Computers and Games}, pp.\  72--83, Berlin, Heidelberg,
  2007. Springer Berlin Heidelberg.
\newblock ISBN 978-3-540-75538-8.

\bibitem[Deng et~al.(2025)Deng, Da, Pan, Sun, Wu, Bhutani,
  et~al.]{deng2025swebenchpro}
Xiang Deng, Jeff Da, Edwin Pan, Yannis~Yiming Sun, Chen Bo~Calvin Wu, Sanyam
  Bhutani, et~al.
\newblock {SWE-Bench Pro}: Can {AI} agents solve long-horizon software
  engineering tasks?, 2025.
\newblock URL \url{https://arxiv.org/abs/2509.16941}.

\bibitem[Dong et~al.(2026)Dong, He, Liu, Hou, Du, Xu, Yu, Yang, Xia, and
  Chen]{dong2026deltabox}
Yunpeng Dong, Jingkai He, Shiqi Liu, Yuze Hou, Dong Du, Zhonghu Xu, Si~Yu,
  Baochuan Yang, Yubin Xia, and Haibo Chen.
\newblock {DeltaBox}: Scaling stateful {AI} agents with millisecond-level
  sandbox checkpoint/rollback.
\newblock \emph{arXiv preprint arXiv:2605.22781}, 2026.
\newblock URL \url{https://arxiv.org/abs/2605.22781}.

\bibitem[Ehrlich et~al.(2025)Ehrlich, Brown, Juravsky, Clark, R{\'e}, and
  Mirhoseini]{ehrlich2025codemonkeys}
Ryan Ehrlich, Bradley Brown, Jordan Juravsky, Ronald Clark, Christopher R{\'e},
  and Azalia Mirhoseini.
\newblock {CodeMonkeys}: Scaling test-time compute for software engineering.
\newblock \emph{arXiv preprint arXiv:2501.14723}, 2025.
\newblock URL \url{https://arxiv.org/abs/2501.14723}.

\bibitem[Gou et~al.(2023)Gou, Shao, Gong, Shen, Yang, Duan, and
  Chen]{gou2023critic}
Zhibin Gou, Zhihong Shao, Yeyun Gong, Yelong Shen, Yujiu Yang, Nan Duan, and
  Weizhu Chen.
\newblock {CRITIC}: Large language models can self-correct with
  tool-interactive critiquing, 2023.
\newblock URL \url{https://arxiv.org/abs/2305.11738}.

\bibitem[Guan et~al.(2025)Guan, Zhang, Liu, Shang, Sun, Zhu, Yang, and
  Yang]{guan2025rstar}
Xinyu Guan, Li~Lyna Zhang, Yifei Liu, Ning Shang, Youran Sun, Yi~Zhu, Fan Yang,
  and Mao Yang.
\newblock r{S}tar-math: Small {LLM}s can master math reasoning with
  self-evolved deep thinking.
\newblock In Aarti Singh, Maryam Fazel, Daniel Hsu, Simon Lacoste-Julien, Felix
  Berkenkamp, Tegan Maharaj, Kiri Wagstaff, and Jerry Zhu (eds.),
  \emph{Proceedings of the 42nd International Conference on Machine Learning},
  volume 267 of \emph{Proceedings of Machine Learning Research}, pp.\
  20640--20661. PMLR, 13--19 Jul 2025.
\newblock URL \url{https://proceedings.mlr.press/v267/guan25f.html}.

\bibitem[Hao et~al.(2023)Hao, Gu, Ma, Hong, Wang, Wang, and
  Hu]{hao2023reasoning}
Shibo Hao, Yi~Gu, Haodi Ma, Joshua~Jiahua Hong, Zhen Wang, Daisy~Zhe Wang, and
  Zhiting Hu.
\newblock Reasoning with language model is planning with world model, 2023.
\newblock URL \url{https://arxiv.org/abs/2305.14992}.

\bibitem[{Harbor Framework Team}(2026)]{harbor2026}
{Harbor Framework Team}.
\newblock Harbor: A framework for evaluating and optimizing agents and models
  in container environments, 2026.
\newblock URL \url{https://github.com/harbor-framework/harbor}.

\bibitem[Huang et~al.(2023)Huang, Chen, Mishra, Zheng, Yu, Song, and
  Zhou]{huang2023selfcorrect}
Jie Huang, Xinyun Chen, Swaroop Mishra, Huaixiu~Steven Zheng, Adams~Wei Yu,
  Xinying Song, and Denny Zhou.
\newblock Large language models cannot self-correct reasoning yet, 2023.
\newblock URL \url{https://arxiv.org/abs/2310.01798}.

\bibitem[Huang et~al.(2026)Huang, Srivatsa, Asch, Patwa, and
  Zhang]{huang2026tclone}
Yutong Huang, Vikranth Srivatsa, Alex Asch, Hansin~Tushar Patwa, and Yiying
  Zhang.
\newblock {TClone}: Low-latency forking of live {GUI} environments for
  computer-use agents.
\newblock \emph{arXiv preprint arXiv:2605.17320}, 2026.
\newblock URL \url{https://arxiv.org/abs/2605.17320}.

\bibitem[Jain et~al.(2025)Jain, Singh, Shetty, Zheng, Sen, and
  Stoica]{jain2025r2egym}
Naman Jain, Jaskirat Singh, Manish Shetty, Liang Zheng, Koushik Sen, and Ion
  Stoica.
\newblock {R2E-Gym}: Procedural environments and hybrid verifiers for scaling
  open-weights {SWE} agents.
\newblock In \emph{Conference on Language Modeling (COLM)}, 2025.
\newblock URL \url{https://arxiv.org/abs/2504.07164}.

\bibitem[Kocsis \& Szepesv{\'a}ri(2006)Kocsis and
  Szepesv{\'a}ri]{kocsis2006uct}
Levente Kocsis and Csaba Szepesv{\'a}ri.
\newblock Bandit based {Monte-Carlo} planning.
\newblock In \emph{Machine Learning: ECML 2006}, volume 4212 of \emph{Lecture
  Notes in Computer Science}, pp.\  282--293. Springer, 2006.

\bibitem[Koh et~al.(2024)Koh, McAleer, Fried, and
  Salakhutdinov]{koh2024treesearch}
Jing~Yu Koh, Stephen McAleer, Daniel Fried, and Ruslan Salakhutdinov.
\newblock Tree search for language model agents.
\newblock \emph{arXiv preprint arXiv:2407.01476}, 2024.
\newblock URL \url{https://arxiv.org/abs/2407.01476}.

\bibitem[Kwok et~al.(2026)Kwok, Li, Atreya, Liu, Jiang, Finn, Pavone, Stoica,
  and Mirhoseini]{kwok2026verifier}
Jacky Kwok, Shulu Li, Pranav Atreya, Yuejiang Liu, Yixing Jiang, Chelsea Finn,
  Marco Pavone, Ion Stoica, and Azalia Mirhoseini.
\newblock {LLM-as-a-Verifier}: A general-purpose verification framework, 2026.
\newblock URL \url{https://arxiv.org/abs/2607.05391}.

\bibitem[Li et~al.(2025)Li, Cao, Cao, Li, Tan, Keutzer, Xing, Gonzalez, and
  Stoica]{li2025sstar}
Dacheng Li, Shiyi Cao, Chengkun Cao, Xiuyu Li, Shangyin Tan, Kurt Keutzer,
  Jiarong Xing, Joseph~E. Gonzalez, and Ion Stoica.
\newblock S*: Test time scaling for code generation, 2025.
\newblock URL \url{https://arxiv.org/abs/2502.14382}.

\bibitem[Li et~al.(2022)Li, Choi, Chung, Kushman, Schrittwieser, Leblond,
  Eccles, Keeling, Gimeno, Dal~Lago, Hubert, Choy, de~Masson~d'Autume,
  Babuschkin, Chen, Huang, Welbl, Gowal, Cherepanov, Molloy, Mankowitz,
  Sutherland~Robson, Kohli, de~Freitas, Kavukcuoglu, and
  Vinyals]{li2022competition}
Yujia Li, David Choi, Junyoung Chung, Nate Kushman, Julian Schrittwieser,
  R{\'e}mi Leblond, Tom Eccles, James Keeling, Felix Gimeno, Agustin Dal~Lago,
  Thomas Hubert, Peter Choy, Cyprien de~Masson~d'Autume, Igor Babuschkin,
  Xinyun Chen, Po-Sen Huang, Johannes Welbl, Sven Gowal, Alexey Cherepanov,
  James Molloy, Daniel~J. Mankowitz, Esme Sutherland~Robson, Pushmeet Kohli,
  Nando de~Freitas, Koray Kavukcuoglu, and Oriol Vinyals.
\newblock Competition-level code generation with {AlphaCode}.
\newblock \emph{Science}, 378\penalty0 (6624):\penalty0 1092--1097, 2022.
\newblock \doi{10.1126/science.abq1158}.
\newblock URL \url{https://doi.org/10.1126/science.abq1158}.

\bibitem[Lightman et~al.(2024)Lightman, Kosaraju, Burda, Edwards, Baker, Lee,
  Leike, Schulman, Sutskever, and Cobbe]{lightman2024lets}
Hunter Lightman, Vineet Kosaraju, Yuri Burda, Harrison Edwards, Bowen Baker,
  Teddy Lee, Jan Leike, John Schulman, Ilya Sutskever, and Karl Cobbe.
\newblock Let's verify step by step.
\newblock In \emph{The Twelfth International Conference on Learning
  Representations}, 2024.
\newblock URL
  \url{https://proceedings.iclr.cc/paper_files/paper/2024/file/aca97732e30bcf1303bc22ac3924fd16-Paper-Conference.pdf}.

\bibitem[Lu et~al.(2025)Lu, Holleis, Zhang, Aumayer, Nan, Bai, Ma, Ma, Li, Yin,
  Wang, and Pang]{lu2025toolsandbox}
Jiarui Lu, Thomas Holleis, Yizhe Zhang, Bernhard Aumayer, Feng Nan, Haoping
  Bai, Shuang Ma, Shen Ma, Mengyu Li, Guoli Yin, Zirui Wang, and Ruoming Pang.
\newblock {T}ool{S}andbox: A stateful, conversational, interactive evaluation
  benchmark for {LLM} tool use capabilities.
\newblock In Luis Chiruzzo, Alan Ritter, and Lu~Wang (eds.), \emph{Findings of
  the Association for Computational Linguistics: NAACL 2025}, pp.\  1160--1183,
  Albuquerque, New Mexico, April 2025. Association for Computational
  Linguistics.
\newblock ISBN 979-8-89176-195-7.
\newblock \doi{10.18653/v1/2025.findings-naacl.65}.
\newblock URL \url{https://aclanthology.org/2025.findings-naacl.65/}.

\bibitem[L{\`u} et~al.(2025)L{\`u}, Kazemnejad, Meade, Patel, Shin, Zambrano,
  Stanczak, Shaw, Pal, and Reddy]{lu2025agentrewardbench}
Xing~Han L{\`u}, Amirhossein Kazemnejad, Nicholas Meade, Arkil Patel, Dongchan
  Shin, Alejandra Zambrano, Karolina Stanczak, Peter Shaw, Christopher Pal, and
  Siva Reddy.
\newblock {AgentRewardBench}: Evaluating automatic evaluations of web agent
  trajectories.
\newblock In \emph{Second Conference on Language Modeling}, 2025.
\newblock URL \url{https://openreview.net/forum?id=fQcUZMPIvu}.

\bibitem[Madaan et~al.(2023)Madaan, Tandon, Gupta, Hallinan, Gao, Wiegreffe,
  Alon, Dziri, Prabhumoye, Yang, Gupta, Majumder, Hermann, Welleck,
  Yazdanbakhsh, and Clark]{madaan2023selfrefine}
Aman Madaan, Niket Tandon, Prakhar Gupta, Skyler Hallinan, Luyu Gao, Sarah
  Wiegreffe, Uri Alon, Nouha Dziri, Shrimai Prabhumoye, Yiming Yang, Shashank
  Gupta, Bodhisattwa~Prasad Majumder, Katherine Hermann, Sean Welleck, Amir
  Yazdanbakhsh, and Peter Clark.
\newblock {Self-Refine}: Iterative refinement with self-feedback, 2023.
\newblock URL \url{https://arxiv.org/abs/2303.17651}.

\bibitem[Marten et~al.(2026)Marten, Shaw, Bercovich, Droste, Cerruti, Dillmann,
  Wang, Wahdany, Hart, Krauth, {ScaleAI}, {Snorkel AI}, {Turing}, {gNucleus
  AI}, {Boolean AI}, Carlini, Lyu, Wei, Khatua, Plüster, Dwivedi, Sutcliffe,
  {Yuming}, Tivris, Wang, Goh, Xing, Lin, Salia, Seol, Bao, Ouyang, Park,
  Walsh, Kong, Ivanov, Ubl, Liamets, Menis, Migdal, Bao, Movva, Ben~Chaim,
  Srinath, Bogdanik, Yadav, Benjamin, Kung, Hughes, Lan, Gupta, Mishra, Wang,
  He, Tu, Montgomery, Tu, Naik, Mortensen, Zhang, Mathur, Liu, Singh, Yu, Feng,
  Gangal, Tao, Ruan, Mueller, Cabezas, Bauer, Li, Zhang, Feller, Madayan, Chen,
  Feuer, Li, Li, Raj, Galler, Shi, Segal, Buchanan, P., Desai, Schneider,
  Settles, Lin, Nezhurina, Wang, Kowalczyk, Zhao, Satia, Hu, Atef, Chen, Vance,
  Segato, Jitsev, Dimakis, Merrill, Konwinski, and
  Schmidt]{Marten_Terminal-Bench_2026}
Ryan Marten, Alex Shaw, Ivan Bercovich, Benedikt Droste, Tommaso Cerruti,
  Steven Dillmann, Ruiyang Wang, Dariush Wahdany, Allen Hart, Karl Krauth,
  {ScaleAI}, {Snorkel AI}, {Turing}, {gNucleus AI}, {Boolean AI}, Nicholas
  Carlini, Shengrui Lyu, Anjiang Wei, Arpandeep Khatua, Björn Plüster,
  Chaitanya Dwivedi, Christine Sutcliffe, {Yuming}, David Tivris, Di~Wang,
  Hui~Wen Goh, Hanwen Xing, Haowei Lin, Irakli Salia, Jaejung Seol, Jiajun Bao,
  Jialin Ouyang, Junha Park, Liam Walsh, Luyang Kong, Maksim Ivanov, Malte Ubl,
  Mikhail Liamets, Orfeas Menis, Piotr Migdal, Qingquan Bao, Raj Movva, Roey
  Ben~Chaim, Namburi Srinath, Sergey Bogdanik, Shubham Yadav, Stephen Benjamin,
  Tony Kung, Walker Hughes, Xin Lan, Himanshu Gupta, Swaroop Mishra, Chenguang
  Wang, Hao He, Jianhong Tu, Kyle Montgomery, Zengji Tu, Atharva Naik, David
  Mortensen, Ivan Zhang, Yash Mathur, Emmy Liu, Karanpartap Singh, Michael Yu,
  Steven Feng, Varun Gangal, Zhuofu Tao, Sherry Ruan, Jonas Mueller, Joan
  Cabezas, Justin Bauer, Kevin~Xiang Li, Robert Zhang, Aaron Feller, Alec
  Madayan, Leon Chen, Ben Feuer, Xiangyi Li, Boxuan Li, Harsh Raj, Samuel
  Galler, Lin Shi, Ivgeni Segal, Kelly Buchanan, Shreyas P., Rishi Desai, Aaron
  Schneider, Chris Settles, Xiangning Lin, Marianna Nezhurina, Andrew Wang,
  Michał Kowalczyk, Jin-Xiang Zhao, Sanyam Satia, Jessie Hu, Sherif Atef, Kobe
  Chen, Sam Vance, Gian Segato, Jenia Jitsev, Alex Dimakis, Mike Merrill, Andy
  Konwinski, and Ludwig Schmidt.
\newblock {Terminal-Bench}, August 2026.
\newblock URL \url{https://github.com/harbor-framework/terminal-bench}.

\bibitem[Shi et~al.(2026)Shi, Wang, Zhao, Chen, Feng, Hao, Su, GU, Su, Cai, and
  He]{shi2026ajbench}
Wentao Shi, Yu~Wang, Yuyang Zhao, Yuxin Chen, Fuli Feng, Xueyuan Hao, Xi~Su,
  Qi~GU, Hui Su, Xunliang Cai, and Xiangnan He.
\newblock {AJ}-bench: Benchmarking agent-as-a-judge for environment-aware
  evaluation.
\newblock In Maria Liakata, Viviane~P. Moreira, Jiajun Zhang, and David Jurgens
  (eds.), \emph{Findings of the {A}ssociation for {C}omputational
  {L}inguistics: {ACL} 2026}, pp.\  25371--25413, San Diego, California, United
  States, July 2026. Association for Computational Linguistics.
\newblock ISBN 979-8-89176-395-1.
\newblock \doi{10.18653/v1/2026.findings-acl.1269}.
\newblock URL \url{https://aclanthology.org/2026.findings-acl.1269/}.

\bibitem[Shinn et~al.(2023)Shinn, Cassano, Gopinath, Narasimhan, and
  Yao]{shinn2023reflexion}
Noah Shinn, Federico Cassano, Ashwin Gopinath, Karthik Narasimhan, and Shunyu
  Yao.
\newblock Reflexion: Language agents with verbal reinforcement learning.
\newblock In \emph{Advances in Neural Information Processing Systems},
  volume~36. Curran Associates, Inc., 2023.

\bibitem[Shum et~al.(2025)Shum, Hui, Chen, Zhang, W., Yang, Huang, Lin, and
  He]{shum2025swerm}
KaShun Shum, Binyuan Hui, Jiawei Chen, Lei Zhang, X.~W., Jiaxi Yang, Yuzhen
  Huang, Junyang Lin, and Junxian He.
\newblock {SWE-RM}: Execution-free feedback for software engineering agents.
\newblock \emph{arXiv preprint arXiv:2512.21919}, 2025.
\newblock URL \url{https://arxiv.org/abs/2512.21919}.

\bibitem[Silver et~al.(2016)Silver, Huang, Maddison, Guez, Sifre, van~den
  Driessche, Schrittwieser, Antonoglou, Panneershelvam, Lanctot, Dieleman,
  Grewe, Nham, Kalchbrenner, Sutskever, Lillicrap, Leach, Kavukcuoglu, Graepel,
  and Hassabis]{silver2016mastering}
David Silver, Aja Huang, Chris~J. Maddison, Arthur Guez, Laurent Sifre, George
  van~den Driessche, Julian Schrittwieser, Ioannis Antonoglou, Veda
  Panneershelvam, Marc Lanctot, Sander Dieleman, Dominik Grewe, John Nham, Nal
  Kalchbrenner, Ilya Sutskever, Timothy Lillicrap, Madeleine Leach, Koray
  Kavukcuoglu, Thore Graepel, and Demis Hassabis.
\newblock Mastering the game of {Go} with deep neural networks and tree search.
\newblock \emph{Nature}, 529\penalty0 (7587):\penalty0 484--489, 2016.
\newblock \doi{10.1038/nature16961}.

\bibitem[Silver et~al.(2018)Silver, Hubert, Schrittwieser, Antonoglou, Lai,
  Guez, Lanctot, Sifre, Kumaran, Graepel, Lillicrap, Simonyan, and
  Hassabis]{silver2018general}
David Silver, Thomas Hubert, Julian Schrittwieser, Ioannis Antonoglou, Matthew
  Lai, Arthur Guez, Marc Lanctot, Laurent Sifre, Dharshan Kumaran, Thore
  Graepel, Timothy Lillicrap, Karen Simonyan, and Demis Hassabis.
\newblock A general reinforcement learning algorithm that masters chess, shogi,
  and {Go} through self-play.
\newblock \emph{Science}, 362\penalty0 (6419):\penalty0 1140--1144, 2018.
\newblock \doi{10.1126/science.aar6404}.

\bibitem[Stroebl et~al.(2026)Stroebl, Kapoor, and Narayanan]{stroebl2026limits}
Benedikt Stroebl, Sayash Kapoor, and Arvind Narayanan.
\newblock The limits of inference scaling through resampling.
\newblock In \emph{The Fourteenth International Conference on Learning
  Representations}, 2026.
\newblock URL \url{https://openreview.net/forum?id=j8H84v6AZ1}.

\bibitem[Trivedi et~al.(2024)Trivedi, Khot, Hartmann, Manku, Dong, Li, Gupta,
  Sabharwal, and Balasubramanian]{trivedi2024appworld}
Harsh Trivedi, Tushar Khot, Mareike Hartmann, Ruskin Manku, Vinty Dong, Edward
  Li, Shashank Gupta, Ashish Sabharwal, and Niranjan Balasubramanian.
\newblock {AppWorld}: A controllable world of apps and people for benchmarking
  interactive coding agents.
\newblock In \emph{Proceedings of the 62nd Annual Meeting of the Association
  for Computational Linguistics (Volume 1: Long Papers)}. ACL, 2024.
\newblock URL \url{https://aclanthology.org/2024.acl-long.850/}.

\bibitem[Wang et~al.(2024)Wang, Li, Shao, Xu, Dai, Li, Chen, Wu, and
  Sui]{wang2024mathshepherd}
Peiyi Wang, Lei Li, Zhihong Shao, Runxin Xu, Damai Dai, Yifei Li, Deli Chen,
  Yu~Wu, and Zhifang Sui.
\newblock {Math-Shepherd}: Verify and reinforce {LLM}s step-by-step without
  human annotations.
\newblock In Lun-Wei Ku, Andre Martins, and Vivek Srikumar (eds.),
  \emph{Proceedings of the 62nd Annual Meeting of the Association for
  Computational Linguistics (Volume 1: Long Papers)}, pp.\  9426--9439,
  Bangkok, Thailand, August 2024. Association for Computational Linguistics.
\newblock \doi{10.18653/v1/2024.acl-long.510}.
\newblock URL \url{https://aclanthology.org/2024.acl-long.510/}.

\bibitem[Wang et~al.(2023)Wang, Wei, Schuurmans, Le, Chi, Narang, Chowdhery,
  and Zhou]{wang2023selfconsistency}
Xuezhi Wang, Jason Wei, Dale Schuurmans, Quoc~V Le, Ed~H. Chi, Sharan Narang,
  Aakanksha Chowdhery, and Denny Zhou.
\newblock Self-consistency improves chain of thought reasoning in language
  models.
\newblock In \emph{The Eleventh International Conference on Learning
  Representations}, 2023.
\newblock URL \url{https://openreview.net/forum?id=1PL1NIMMrw}.

\bibitem[Wu et~al.(2026)Wu, Chang, Cao, Gao, and Wang]{wu2026crab}
Tianyuan Wu, Chaokun Chang, Lunxi Cao, Wei Gao, and Wei Wang.
\newblock Crab: A semantics-aware checkpoint/restore runtime for agent
  sandboxes.
\newblock \emph{arXiv preprint arXiv:2604.28138}, 2026.
\newblock URL \url{https://arxiv.org/abs/2604.28138}.

\bibitem[Xie et~al.(2023)Xie, Kawaguchi, Zhao, Zhao, Kan, He, and
  Xie]{xie2023self}
Yuxi Xie, Kenji Kawaguchi, Yiran Zhao, James~Xu Zhao, Min-Yen Kan, Junxian He,
  and Michael Xie.
\newblock Self-evaluation guided beam search for reasoning.
\newblock In A.~Oh, T.~Naumann, A.~Globerson, K.~Saenko, M.~Hardt, and
  S.~Levine (eds.), \emph{Advances in Neural Information Processing Systems},
  volume~36, pp.\  41618--41650. Curran Associates, Inc., 2023.
\newblock \doi{10.52202/075280-1802}.
\newblock URL
  \url{https://proceedings.neurips.cc/paper_files/paper/2023/file/81fde95c4dc79188a69ce5b24d63010b-Paper-Conference.pdf}.

\bibitem[Xu et~al.(2025)Xu, Zhou, Wu, and Kaffes]{xu2025foundations}
Jiakai Xu, Tianle Zhou, Eugene Wu, and Kostis Kaffes.
\newblock Toward systems foundations for agentic exploration.
\newblock \emph{arXiv preprint arXiv:2510.05556}, 2025.
\newblock URL \url{https://arxiv.org/abs/2510.05556}.

\bibitem[Xu et~al.(2026)Xu, Ge, Timmaraju, Liargkovas, Wu, and
  Kaffes]{Xu_Waypoint_2026}
Jiakai Xu, Tiancheng Ge, Rohan Timmaraju, Georgios Liargkovas, Eugene Wu, and
  Kostis Kaffes.
\newblock {Waypoint}, September 2026.
\newblock URL \url{https://github.com/Alex-XJK/waypoint}.

\bibitem[Yao et~al.(2023)Yao, Yu, Zhao, Shafran, Griffiths, Cao, and
  Narasimhan]{yao2023tree}
Shunyu Yao, Dian Yu, Jeffrey Zhao, Izhak Shafran, Tom Griffiths, Yuan Cao, and
  Karthik Narasimhan.
\newblock Tree of thoughts: Deliberate problem solving with large language
  models.
\newblock In A.~Oh, T.~Naumann, A.~Globerson, K.~Saenko, M.~Hardt, and
  S.~Levine (eds.), \emph{Advances in Neural Information Processing Systems},
  volume~36, pp.\  11809--11822. Curran Associates, Inc., 2023.
\newblock \doi{10.52202/075280-0517}.
\newblock URL
  \url{https://proceedings.neurips.cc/paper_files/paper/2023/file/271db9922b8d1f4dd7aaef84ed5ac703-Paper-Conference.pdf}.

\bibitem[Yu et~al.(2025)Yu, Li, and Wang]{yu2025scaling}
Fei Yu, Yingru Li, and Benyou Wang.
\newblock Scaling flaws of verifier-guided search in mathematical reasoning,
  2025.
\newblock URL \url{https://arxiv.org/abs/2502.00271}.

\bibitem[Zhou et~al.(2024)Zhou, Yan, Shlapentokh-Rothman, Wang, and
  Wang]{zhou2024lats}
Andy Zhou, Kai Yan, Michal Shlapentokh-Rothman, Haohan Wang, and Yu-Xiong Wang.
\newblock Language agent tree search unifies reasoning, acting, and planning in
  language models.
\newblock In \emph{Proceedings of the 41st International Conference on Machine
  Learning (ICML)}, volume 235 of \emph{PMLR}, 2024.
\newblock URL \url{https://proceedings.mlr.press/v235/zhou24r.html}.

\bibitem[Zhuge et~al.(2024)Zhuge, Zhao, Ashley, Wang, Khizbullin, Xiong, Liu,
  Chang, Krishnamoorthi, Tian, Shi, Chandra, and
  Schmidhuber]{zhuge2024agentjudge}
Mingchen Zhuge, Changsheng Zhao, Dylan Ashley, Wenyi Wang, Dmitrii Khizbullin,
  Yunyang Xiong, Zechun Liu, Ernie Chang, Raghuraman Krishnamoorthi, Yuandong
  Tian, Yangyang Shi, Vikas Chandra, and J{\"u}rgen Schmidhuber.
\newblock Agent-as-a-judge: Evaluate agents with agents.
\newblock \emph{arXiv preprint arXiv:2410.10934}, 2024.
\newblock URL \url{https://arxiv.org/abs/2410.10934}.

\end{thebibliography}
